\documentclass[11pt]{article}
\usepackage[final]{acl}
\usepackage{times}
\usepackage{latexsym}
\usepackage[T1]{fontenc}
\usepackage[utf8]{inputenc}
\usepackage{microtype}
\usepackage{inconsolata}
\usepackage{graphicx}
\usepackage{graphicx}
\usepackage{xcolor}
\usepackage{xspace}
\usepackage{dblfloatfix}
\usepackage{latexsym}
\usepackage{algorithm}
\usepackage{algorithmic}
\usepackage{amsmath}
\usepackage{multirow}
\usepackage{listings}
\usepackage{makecell}
\usepackage{siunitx}
\usepackage{booktabs}

\usepackage{tcolorbox}
\tcbuselibrary{breakable, skins}
 
\def\myalgo{AdaBPE\xspace}

\def\myalgolong{Adaptive Frequency Threshold Byte Pair Encoding\xspace} 
\def\myalgoshort{Adaptive Frequency Threshold BPE\xspace} 

\title{Adaptive Frequency Threshold Byte Pair Encoding for Mixed Natural Language and Structured Symbolic Corpora}

\author{Pietro Di Nicco, Lucio La Cava, Andrea Tagarelli \\
  DIMES Department, University of Calabria, Italy \\
  \texttt{\{lucio.lacava, tagarelli\}@dimes.unical.it}}

\begin{document}
\maketitle
\begin{abstract} 
Modern language models rely on subwording tokenization to define the discrete units of input processing, typically using methods like Byte Pair Encoding (BPE) that treat whitespace as a fixed semantic boundary. This design is inherently limited, as many meaningful concepts---including multiword expressions and the structured environments of formal languages like \LaTeX{}---frequently span across whitespaces. While recent superword tokenization methods attempt to address this by relaxing whitespace constraints, they often rely on greedy frequency-driven merging that fails to account for token length or the specific needs of underrepresented domain-specific languages.
In this paper, we introduce \myalgolong  (\myalgo), a new tokenization method that leverages an adaptive frequent-pattern mining approach. \myalgo features a twofold dynamicity: it adapts minimum frequency thresholds based on both the length of a candidate token, to prevent favoring frequent short patterns, and its type, to distinguish between general and formal language patterns, using \LaTeX{} as case in point. This allows the model to efficiently allocate vocabulary capacity to domain-specific notations, such as \LaTeX{}, even when they represent a small fraction of the training data. 
We evaluate \myalgo across five datasets spanning \LaTeX{} and natural language, showing consistent gains in encoding efficiency and downstream language-model performance on domain-specific content without compromising general-language performance.
\end{abstract}

\section{Introduction}
In modern language models, the goal of tokenization is to construct a vocabulary of atomic units, called \textit{tokens}, from a large text corpus. This vocabulary defines the discrete symbols that the model can process as input. 
The main approach for learning such a vocabulary is \emph{subword tokenization}. Widely used algorithms, e.g., Byte Pair Encoding (BPE) \cite{gage1994bpe}, WordPiece \cite{schuster2012japanese}, and Unigram \cite{subword_kudo_2018}, learn a vocabulary of subword units from the training corpus. These methods typically treat whitespace as a semantic boundary between words, using it to constrain how text is segmented into subword tokens. 

This design however has an inherent limitation: whitespace is not, in general, a reliable indicator of semantic boundaries. Many concepts are expressed as \textit{multiword expressions}, such as ``Artificial Intelligence'' or ``state of the art'', where semantically meaningful units often span multiple whitespace-separated tokens. 
Ideally, such expressions should be represented as a single lexical unit during tokenization. 
A similar issue arises in \textit{formal languages} including programming languages and markup, such  as HTML, Markdown, or \LaTeX{}. 
  For example, the expression \texttt{\textbackslash begin\{equation\}} $\ldots$ \texttt{\textbackslash end\{equation\}} denotes a single mathematical environment, yet conventional subword tokenizers process each command, delimiter, and surrounding text separately.

\vspace{1mm}
\noindent
\textbf{Related work.}  
Recent studies have revisited the objectives of conventional subword tokenization  approaches, highlighting their inherent limitations.  
\citet{lim2025partition} frame   subword tokenization   as an NP-hard problem, and propose a greedy alternative called GreedTok. 
Similarly, \citet{asgari-etal-2026-morphbpe} propose MorphBPE to  constrain  BPE merge operations to respect morphene boundaries, improving morphological consistency. 
\citet{phan2024tokenization} relate   subword tokenization   to a bias phenomenon,
while \cite{zhang2024tokenization} investigated  negative impacts on the   LLM performance on   reasoning tasks.

A complementary body of research has focused on vocabulary design.  \citet{tao2024scaling} analyze   the compute-optimal vocabulary size growing  with the model scale, whereas \citet{lai-etal-2021-lattice,kumar-thawani-2022-bpe,zhang-etal-2021-ambert} explore  incorporating larger linguistic units into the vocabulary; however, these approaches are limited to specific tasks (e.g., machine translation) or languages.  
More recently, a new trend of \textit{``superword'' tokenization} has emerged. Particularly,  SuperBPE \cite{liu-etal-2025-superbpe} and BoundlessBPE \cite{schmidt2025boundless}   adopt a variant of BPE that, after initialization like standard BPE, relaxes the pre-tokenization constraint allowing merge candidates that cross whitespace boundaries. The obtained results highlight the benefits of multiword tokens in improving compression, reducing training costs, and accelerating inference.

\vspace{1mm}
\noindent
\textbf{Motivations of this study.} 
However, two further issues  remain largely unaddressed, which are discussed as follows. 
\textbf{(I1)} Although methods like SuperBPE relaxes the constraint that token merges must respect whitespace boundaries, their vocabulary induction procedure remains based on greedy frequency-driven merging. This  greedy merge criterion is based primarily on the frequency of adjacent token pairs rather than directly optimizing the reduction in encoding length. Consequently, it does not explicitly account for the length of candidate tokens or the compression benefit obtained from introducing longer tokens, potentially favoring frequent short patterns over less frequent candidates that could yield larger reductions in sequence length.
 \textbf{(I2)} The above issue is further exacerbated for  domain-specific languages that, in the training data of modern LLMs, is relatively less  represented than other sublanguages or formal languages. A particularly relevant example is the representation of mathematical notation through  \LaTeX{}, which has been shown to improve the reasoning ability of LLMs on mathematical tasks \cite{Taylor2022GalacticaAL}, \cite{Lewkowycz2022SolvingQR}, \cite{Ahn2024LargeLM}.  
Unlike lightweight markup   such as Markdown, \LaTeX{} encodes mathematical content through a rich hierarchy of commands and environments  which   preserves syntactic and semantic structure that is essential for mathematical reasoning, suggesting that \LaTeX{} constitutes a more informative training signal for mathematical language modeling than markup languages designed primarily for document formatting.
  \LaTeX{} commands and environments define structured semantic units that are largely independent of whitespace, 
highlighting the mismatch between whitespace-based tokenization and the underlying semantic structure of the document.

\vspace{1mm}
\noindent\textbf{Contributions.}
In this paper, we aim to address the above issues (I1-I2) by introducing    a new tokenization method, named \textit{\myalgo}, which leverages an \textit{adaptive frequent-pattern mining approach} to identify multiword expressions  and build an enhanced BPE-based  vocabulary. 
\myalgo is designed to explicitly distinguish general language patterns from formal language patterns, with emphasis on  \LaTeX{}, in inducing supertokens.
The key feature of \myalgo is a twofold dynamicity in determining token merges, since it adapts the  minimum frequency  threshold of a  token pattern to both its \textit{length} (cf. I1) and its  \textit{type}---thus   differently treating  frequent multiword expressions belonging to either  \LaTeX{}   or general language (cf. I2). 
Yet, \myalgo provides  a single vocabulary-construction procedure, which  allocates capacity differently to general natural language and to domain-specific notation, without requiring separate vocabularies. 
This leads to an efficient mechanism for capturing domain-specific languages like \LaTeX\, even if these languages account for only a small fraction of the training data.

Across five datasets spanning both \LaTeX{} and general-language domains, we compared \myalgo against SuperBPE and BPE along four complementary axes---encoding efficiency, supertoken contribution, token-distribution statistics---showing that \myalgo achieves consistent compression gains on domain-specific content without degrading performance on general language. 
Moreover, we trained small-scale language models using \myalgo and SuperBPE tokenizers, and evaluated these models  on five downstream benchmarks to assess the impact  of \myalgo's tokenization strategy on downstream performance.

\section{Methodology}

\subsection{Preliminaries}
\vspace{1mm}
\noindent
\textbf{BPE}~\cite{gage1994bpe} constructs a vocabulary of subword units by iteratively merging the most frequent pair of adjacent tokens in a training corpus. Canonically, 
BPE is applied after whitespace pre-tokenization, restricting merges to occur only within whitespace-delimited words. 
 This design prevents learned units from spanning word boundaries, making BPE effective for capturing morphological structure while limiting its ability to represent frequent multiword expressions as single lexical or semantic units.

\vspace{1mm}
\noindent
\textbf{SuperBPE}~\cite{liu-etal-2025-superbpe} addresses the BPE limitation by introducing a two-stage training procedure:
\textit{Stage 1 (Subword Learning)}: The tokenizer performs standard BPE training up to a transition point, using whitespace pre-tokenization to establish a base vocabulary of   subwords.
\textit{Stage 2 (Supertoken Learning)}: Training resumes from the vocabulary learned in Stage 1, but with whitespace pre-tokenization disabled. By allowing merges to bridge whitespace boundaries, the algorithm encodes common multiword sequences, or ``supertokens'' into the final vocabulary, enhancing tokenization efficiency for frequently occurring patterns.

\subsection{\myalgoshort}
Algorithm~\ref{alg:1} provides an outline of \myalgo, which  takes as input a training corpus $\mathcal{C}$, a BPE vocabulary $V$,  a global threshold $\mathtt{min}\_\mathtt{thr}$, and a maximum number of iterations $I$. 
The key hyperparameter is the threshold $\mathtt{min}\_\mathtt{thr}$, which sets the baseline token-selectivity of \myalgo, and is in turn  modulated by two additional, pattern-level language-dependent hyperparameters, dubbed  $\theta_\mathtt{latex}^\mathtt{min}$ and $\theta_\mathtt{general}^\mathtt{min}$, which serve to adapt the effective threshold to the type of pattern under consideration during the tokenization process.

\begin{algorithm}[t!]
\small
\caption{\myalgoshort (\myalgo)}
\begin{algorithmic}[1]
\label{alg:1}

\STATE \textbf{Input:} corpus $\mathcal{C}$, BPE vocabulary $V$ over $\mathcal{C}$, global threshold $\mathtt{min\_thr} \in [0,1]$,  pattern-specific params. $\theta_\mathtt{general}^\mathtt{min}$ and $\theta_\mathtt{latex}^\mathtt{min}$, and  max iterations $I$

\STATE $i \gets 0$ 
\WHILE{$V$ has changed \textbf{and} $i < I$}
    \STATE $\mathcal{C}_i \gets \textsf{tokenize}(\mathcal{C}, V)$ 
    \STATE $B \gets \textsf{extract\_bigrams}(\mathcal{C}_i)$ 
    \FORALL{$b \in B$}
        \IF{$b$ contains LaTeX or alphanumeric patterns}
            \STATE $\mathtt{min}_b^{\text{latex}} \gets 
                \mathtt{min\_thr} \cdot (\theta_{\text{latex}}^{\min})^{\log_2(nchars(b))}$

            \IF{$\mathtt{min}_b^{\text{latex}} \leq sup(b,\mathcal{C}_i)$} \STATE $V \gets V \cup \{b\}$
            \ENDIF
            
        \hspace{-5mm}\COMMENT{\textit{general language}}
        \ELSE 
            \STATE $\mathtt{min}_b^{\text{general}} \gets 
                \mathtt{min\_thr} \cdot (\theta_{\text{general}}^{\min})^{nspaces(b)+1}$
            
            \IF{$\mathtt{min}_b^{\text{general}} \leq sup(b,\mathcal{C}_i)$}
            \STATE $V \gets V \cup \{b\}$
            \ENDIF
        
        \ENDIF

    \ENDFOR
    \STATE $i \gets i + 1$ 
\ENDWHILE

\end{algorithmic}
\end{algorithm}

At each iteration, the corpus $\mathcal{C}$ is tokenized  based on  whitespace delimitation of currently learned (super)tokens (Line 4). The vocabulary $V$   is initialized with the BPE vocabulary and progressively expanded by adding bigrams identified in previous iterations through frequent-pattern mining.
For each candidate pattern (bigram) $b$ (Lines 5-6), the algorithm decides whether $b$ should be added to $V$ by comparing its support $sup(b,\mathcal{C}_i)= count(b,\mathcal{C}_i)/|\mathcal{C}_i|$ against a frequency threshold, where $count(b,\mathcal{C}_i)$ is the number of occurrences of $b$ in $\mathcal{C}_i$, and $|\mathcal{C}_i|$ is the total number of patterns (i.e., tokens) in the corpus tokenized at iteration $i$. 
If $b$ contains \LaTeX{} content (Lines 7-11), it is added to the vocabulary whenever $sup(b) \geq \mathtt{min}_b^{\mathtt{latex}}$, where $\mathtt{min}_b^{\mathtt{latex}}$ is obtained by scaling $\mathtt{min\_thr}$ with the \LaTeX-specific parameter $\theta_\mathtt{latex}^\mathtt{min}$. Otherwise, if $b$ corresponds to general language (Lines 12-18), its support is compared to a threshold $\mathtt{min}_b^\mathtt{general}$, which  is computed using a general-language parameter $\theta_\mathtt{general}^\mathtt{min}$.  

In determining both $\mathtt{min}_b^\mathtt{general}$ and $\mathtt{min}_b^{\mathtt{latex}}$, 
the length of pattern $b$ is taken into account to control an exponential scaling of $\theta_\mathtt{general}^\mathtt{min}$ and $\theta_\mathtt{latex}^\mathtt{min}$, respectively; for natural language, we consider the number of whitespace-separated words as the length of the pattern, whereas for \LaTeX{}, we consider the base-2 logarithm of the number of characters in the pattern.

\vspace{1mm}
\noindent\textbf{Remarks on frequent pattern mining.}
Since $V$ is updated at every iteration, \myalgo induces an exponential supertoken composition mechanism: at the first iteration, only bigrams derived from BPE tokens are considered; at the second iteration, the bigrams produced in the previous step(s) may themselves be combined, yielding 3-grams and 4-grams w.r.t. the original BPE vocabulary; and so on for longer patterns at subsequent iterations. 
In practice, a very small number of iterations is required to converge---three iterations are sufficient to reach patterns containing up to eight BPE tokens. 

It should also be noted that, fFor any candidate pattern $b$ at iteration $i-1$ that still is at the current iteration $i$, both  $count(b,\mathcal{C}_i)$ and its associated $\mathtt{min}_b$ threshold do not change, however its support is non-decreased since  during \myalgo training, the number of tokens in the corpus decreases as merges are applied.  Consequently, a pattern $b$ could pass the frequency check at subsequent iterations than its first identification. 
Nonetheless, this does not impact  the overall time cost of \myalgo which is bounded by   $O(I \cdot|\mathcal{C}_0|)$, i.e., \myalgo  scales linearly with the size of the training corpus expressed in terms of total number of tokens in the corpus when represented using the initial BPE vocabulary.

\subsubsection{Setting $\mathtt{min\_thr}$.}
To determine the value of the initial threshold $\mathtt{min\_thr}$, we reason about the size of the final vocabulary. We set the final vocabulary size to around 200k tokens, so as to match the one of  SuperBPE. We adopt this value primarily to ensure a controlled comparison across BPE, SuperBPE and \myalgo, thus supporting  our goal of  isolating the effect of the tokenization method. 

\begin{figure}[t!]
    \centering
    \includegraphics[width=7cm,height=4cm]{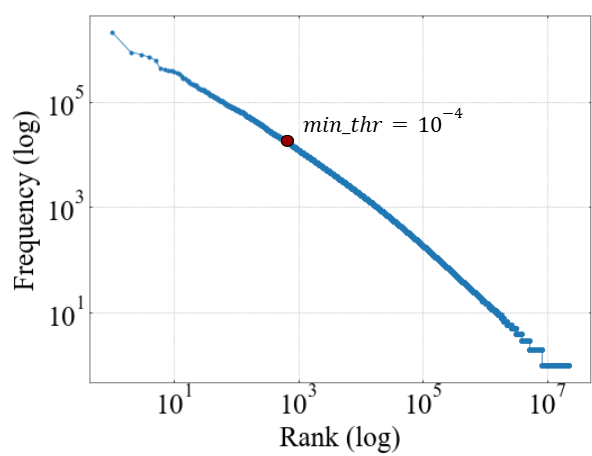}
    \vspace{-1mm}
    \caption{Bigram frequency distribution estimated from a random sample of size $1.5 \cdot 10^8$.\vspace{-5mm}}
    \label{fig:minthr}
\end{figure}

Like in SuperBPE, we set the size of the starting BPE vocabulary to 160k tokens. Since our target final vocabulary is of the order of 200k, \myalgo should add on the order of $10^4$ tokens at each iteration. 
During the first iteration, the algorithm explores the bigram distribution, which is known to follow Zipf's law
\cite{10.2307/2572028}, \cite{bigramlaw}.  We set $\mathtt{min\_thr}$ by reasoning about this first iteration, (Empirically, we have observed that the number of patterns added at each subsequent iteration is of the same order of magnitude as the first one). 
Crucially, $\mathtt{min\_thr}$ represents an \emph{upper bound} on the threshold effectively applied to each pattern, since the language-dependent parameters $\theta_\mathtt{general}^\mathtt{min}, \theta_\mathtt{latex}^\mathtt{min}$ (ranging within $(0,1)$) further reduce it. We hence choose $\mathtt{min\_thr}$ so that only about $10^3$ bigrams have relative support above $\mathtt{min\_thr}$ in the first iteration; the subsequent downward scaling by the language-dependent parameters then raises the number of admitted patterns to the order of $10^4$.   
Figure~\ref{fig:minthr} shows the empirical bigram frequency distribution, estimated from a random sample of $1.5 \cdot 10^8$ bigrams drawn from the training data (see Appendix~\ref{app:training}).
As shown, $\mathtt{min\_thr}=10^{-4}$ is an appropriate choice, since the number of patterns considered in the first iteration remains on the order of $10^3$, before the scaling introduced by the language-dependent parameters.

\subsubsection{Setting $\theta_{\mathtt{general}}^{\mathtt{min}}$ and $\theta_{\mathtt{latex}}^{\mathtt{min}}$.}
\label{sec:hyperparam}

\begin{figure}[t!]
    \centering
   \includegraphics[width=\linewidth]{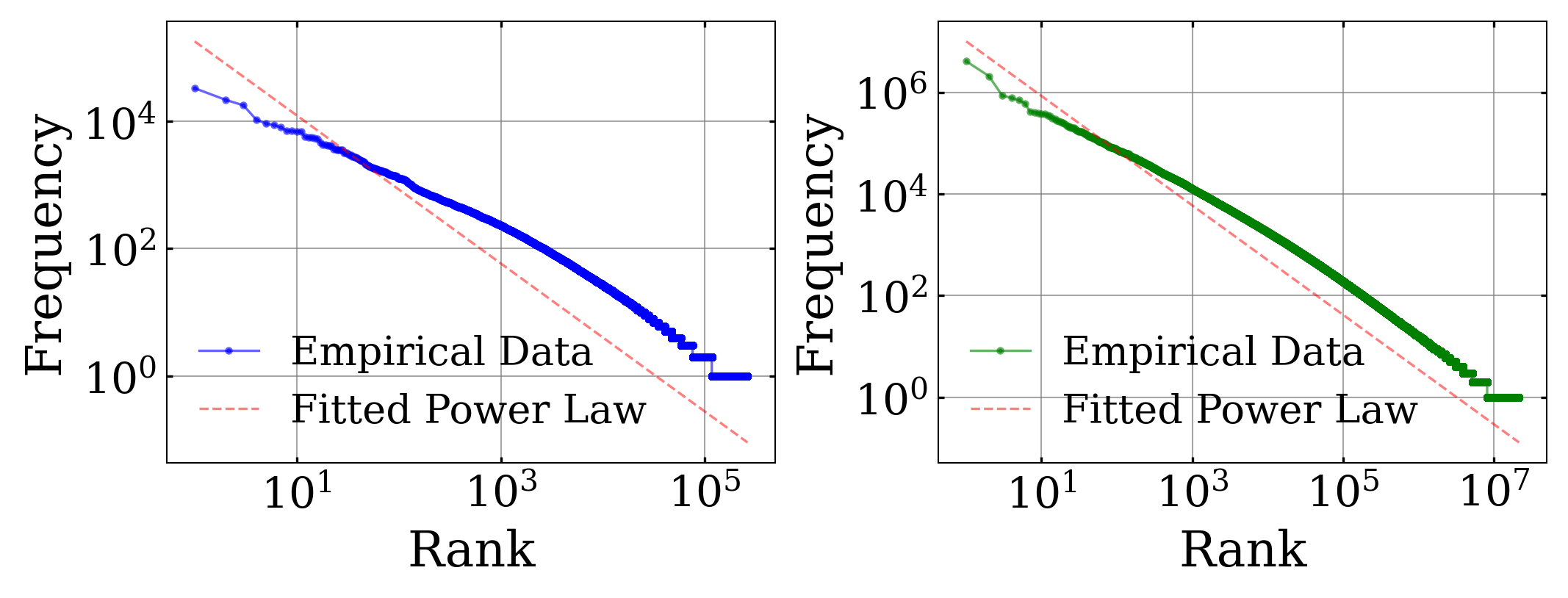}
    \caption {Frequency distribution of natural-language bigrams (right) and of \LaTeX{} bigrams (left).}
    \label{fig:splitdist}
\end{figure}

As  previously mentioned, $\theta_\mathtt{general}^\mathtt{min}$ and $\theta_\mathtt{latex}^\mathtt{min}$ are  language-dependent parameters used to scale the frequency threshold according to the type of pattern. 
Since \LaTeX{} is typically relatively less represented in the training data, the scaling to \LaTeX{} patterns should be more aggressive, therefore we set the constraint   $\theta_\mathtt{general}^\mathtt{min} >\theta_\mathtt{latex}^\mathtt{min}$. 
Following the same approach as for the global threshold, we examined   the frequency distributions of natural-language bigrams and   of \LaTeX{} bigrams (see Figure~\ref{fig:splitdist}). 

To match the conditions set for  the final vocabulary size, in the first iteration the two parameters are bounded based on the number of patterns admitted in the first iteration, i.e., growing the starting 160k BPE vocabulary to the target size
of 200k requires adding $200\text{k}-160\text{k}=4\cdot10^4$ tokens in total. We therefore take $4\cdot10^4$ as the upper bound on the number of patterns admitted for a given language type in the first iteration---admitting more would already exceed, the entire vocabulary budget---and set the lower bound one order of magnitude smaller, at $4\cdot10^3$, so that each language regime contributes a non-negligible number of patterns while leaving headroom for the subsequent iterations: 
$$\small 
 \textsf{sup}\bigl(b_{4 \cdot 10^4}\bigr) \!<\! \mathtt{min\_thr} \cdot \bigl(\theta_{\mathtt{general}}^{\mathtt{min}}\bigr)^{2}
  \!<\! \textsf{sup}\bigl(b_{4 \cdot 10^3}\bigr)
$$
$$\small 
\textsf{sup}\bigl(b_{4 \cdot 10^4}\bigr)\!\!<\! \mathtt{min\_thr} \cdot \bigl(\theta_{\mathtt{latex}}^{\mathtt{min}}\bigr)^{\textsf{nchars}(b)}
 \!\!\!< \!\textsf{sup}\bigl(b_{4 \cdot 10^3}\bigr) 
$$

\noindent 
where $\textsf{sup}(b_x)$ denotes the support of the pattern with rank $x$ in the general-language, resp. \LaTeX{},  distribution.
Since $\mathtt{min\_thr}=10^{-4}$ and  spotting $\textsf{sup}(b_x)$ from  the two empirical distributions (Figure~\ref{fig:splitdist}), we  derive a lower bound and an upper bound for each language-dependent parameter:
$$\small 
3 \cdot 10^{-6} < 10^{-4} \cdot \bigl(\theta_{\mathtt{general}}^{\mathtt{min}}\bigr)^{2} < 3 \cdot 10^{-5}
$$
$$\small 
4 \cdot 10^{-8} < 10^{-4} \cdot \bigl(\theta_{\mathtt{latex}}^{\mathtt{min}}\bigr)^{1.8} < 5 \cdot 10^{-7}
$$

\noindent 
where the exponent $1.8$ for \LaTeX{} patterns is estimated as $\log_2$ of the average pattern length in the \LaTeX{} portion of the training distribution. Solving these inequalities yields
$$\small 
0.2 < \theta_{\mathtt{general}}^{\mathtt{min}} < 0.5,
\qquad
0.02 < \theta_{\mathtt{latex}}^{\mathtt{min}} < 0.06.
$$

\vspace{1mm}
\noindent\textbf{Remarks on hyperparameter tuning.}
We emphasize that the   above settings are not meant as hard constraints but as a
principled way to narrow the search to the correct order of magnitude: they
identify the region in which the number of admitted patterns is compatible with
the target vocabulary size. 
Moreover,  while 
our \myalgo hyperparameter thresholds have been  determined through a data-driven procedure on a representative training sample, these capture general distributional patterns rather than corpus-specific heuristics. This makes the proposed approach robust, transferable, and applicable across diverse datasets.

\subsection{\LaTeX{} and Math Parsing}
To decide whether a candidate pattern should be treated as \LaTeX{}/mathematical content, we test it against the following regular expression:  
\verb!(\\[a-zA-Z]+|[\^_{}]|[a-zA-Z]+[\d_]+|\d+! \verb![a-zA-Z_]+)!.
The expression is a disjunction of four cases, each capturing a syntactic
building block of \LaTeX{} and mathematical notation. The first alternative,
\texttt{\textbackslash\textbackslash[a-zA-Z]+}, matches \LaTeX{} control
sequences, i.e. a backslash followed by a command name (e.g. \verb|\frac|,
\verb|\begin|, \verb|\leftarrow|).
The second, \texttt{[\^{}\_\{\}]},
matches the structural operators of math mode: superscript (\verb|^|), subscript
(\verb|_|), and the grouping braces (\verb|{|, \verb|}|). The last two,
\texttt{[a-zA-Z]+[\textbackslash d\_]+} and
\texttt{\textbackslash d+[a-zA-Z\_]+}, match mixed alphanumeric tokens---letters
followed by digits or underscores and vice versa---which are characteristic of
indexed variables (e.g. \verb|x_|, \verb|theta_|), coefficients, and identifiers found in mathematical/programming notation.
We deliberately favor this lightweight, purely lexical test over a full \LaTeX{}
parser: the check is applied to every candidate bigram during the mining phase,
which runs over a very large corpus, so it must be cheap to evaluate. The
expression is not meant to validate syntactically correct \LaTeX{}, but only to
flag---at scale---the presence of the structural markers that distinguish
domain-specific notation from general  language. 
Following prior work, we tokenize digits from right to left based on the   regular expression \verb|(?=(\d{3})+(?!\d))|, which has been shown to be a suited choice for supporting the arithmetic abilities of LLMs \cite{singh2024tokenization}.

\section{Evaluation Methodology}

\subsection{Experimental Setup}
We evaluated \myalgo against two popular approaches: BPE and SuperBPE. To ensure a fair comparison, all three tokenizers are trained on the same data, which correspond to the SuperBPE training data: the 10\,GB subset of the OLMo-2 stage-2 pretraining corpus (\emph{olmo-mix}) \cite{OLMo20242O2}, more details are given in Appendix~\ref{app:training}.  
  To match the vocabulary size of BPE and SuperBPE (200k), we trained \myalgo for three iterations ({$N\!=\!3$}), with $
\mathtt{min\_thr} = 10^{-4}, \:\:
\theta_{\mathtt{general}}^{\mathtt{min}} = 0.3, \:\:
\theta_{\mathtt{latex}}^{\mathtt{min}} = 0.08$. 
Starting from a BPE vocabulary of 160k tokens, \myalgo reaches a final size of $204{,}601$ tokens.

\subsection{Evaluation Data}
Our evaluation spans five datasets covering the two settings targeted by our method, namely the \textit{technical} and \textit{general} language.  
\textit{latex-formulas} \cite{latex-formulas-dataset} and \textit{Image2Struct} \cite{roberts2024image2struct} are selected to provide a focused test of whether our approach yields the expected benefits on \LaTeX{} based technical documents; 
specifically, 
\textit{latex-formulas} 
consists of mathematical expressions written in \LaTeX{}, 
and \textit{Image2Struct}   contains \LaTeX{} code extracted from arXiv papers. 
Other three datasets---\textit{DROP} \cite{dua-etal-2019-drop}, \textit{SQuAD} \cite{rajpurkar-etal-2016-squad}, and \textit{CommonsenseQA} (CSQA) \cite{talmor-etal-2019-commonsenseqa}---are general natural-language benchmarks, included to verify that this specialization does not come at the cost of compression on   text-heavy documents. 
Summary statistics are reported in Appendix~\ref{app:datasets}.

\subsection{Evaluation Goals and Assessment Criteria}

Our evaluation is guided by a single overarching question: does \myalgo achieve its design goal of better representing under-represented domain-specific languages--- \LaTeX{}---without degrading its treatment of general natural language? 
We address this question through an extensive   evaluation that combines \textit{qualitative} and \textit{quantitative} analyses, comparing \myalgo against BPE and SuperBPE.

\vspace{1mm}
\noindent\textbf{Qualitative analysis.} 
We start by qualitatively examining the merge patterns learned by each tokenizer.
We carry out two complementary analyses: a side-by-side inspection of how the tokenizers segment representative sentences, and a comparison of the most frequent tokens that are exclusive to each tokenizer on every evaluation dataset.

\vspace{2mm}
\noindent  
Our  \textbf{quantitative analysis} is based on three   complementary dimensions:  (i) encoding efficiency, (ii)  contribution of supertokens, and (iii) statistical properties of the  token distribution. 
We next provide intuitive explanations of these criteria; their formal definitions are deferred to Appendix \ref{app:metrics}.

\vspace{1mm}
\noindent
\textbf{Encoding efficiency.}
The most direct measure of tokenizer quality is how compactly it encodes text. We assess this using the \textit{Average Encoding Length} (AEL), defined as the mean number of tokens per document in the corpus. Lower AEL values indicate shorter tokenized sequences, reducing the computational and inference costs of LLMs.

\vspace{1mm}
\noindent
\textbf{Supertoken contribution.}
To characterize the role played by the supertokens, we introduce three measures:   the \emph{Supertoken Share}, i.e., the fraction of emitted token stream made up of supertokens; the \emph{Mean Savings}, i.e., the expected number of base BPE tokens collapsed into a single supertoken; and the \emph{Total Savings}, which combines the two into an overall measure of the compression attributable to supertokens. Together these metrics indicate how heavily each tokenizer relies on supertokens.

\vspace{1mm}
\noindent
\textbf{Token distribution.}
Beyond raw compression, the statistical shape of the token distribution is known to correlate with the quality of a tokenizer for downstream language modeling \cite{zouhar-etal-2023-tokenization}. 
Thus, we measure two criteria:  \emph{N-gram diversity}, i.e.,  the fraction of unique $n$-grams in the encoded text, 
and \emph{R\'enyi efficiency}   \cite{zouhar-etal-2023-tokenization}, which  summarizes how balanced the token-frequency distribution is, where 
higher values indicate a more balanced token-frequency distribution.

\vspace{1mm}
\noindent
\textbf{Downstream task validation.} 
We also assessed the impact of \myalgo vs. SuperBPE downstream behavior. 
 To this aim,  we trained small-scale models  on top of \myalgo and SuperBPE tokenizers, using 
  the full 100GB subset released by \cite{liu-etal-2025-superbpe} as the training data (cf. Appendix~\ref{app:training}) and 
 OLMo-2 \cite{OLMo20242O2} as the decoder architecture. This choice is motivated since OLMo-2 is a \textit{fully open-source}   model with publicly released training data and code, making it well suited for controlled tokenizer comparisons.  
 The two resulting OLMo-2 models, dubbed \textbf{OLMo2-\myalgo} and \textbf{OLMo2-SuperBPE} were trained under identical architecture and budget, differing only in the tokenizer (cf. Appendix~\ref{app:training_details} for hyperparameters).

We resorted to  five evaluation  benchmarks: MATH500 \cite{hendrycksmath2021}, BoolQ \cite{clark2019boolq}, Winogrande \cite{ai2:winogrande}, CSQA \cite{talmor-etal-2019-commonsenseqa} and GSM8K \cite{cobbe2021gsm8k}. 
We measured  the benchmark \textit{accuracy} and  the average \textit{bits-per-byte} (BPB) \cite{gao2020pile800gbdatasetdiverse}; unlike perplexity or other token-level measures, BPB normalizes the model's cross-entropy loss by the number of \emph{bytes} of the original text, making it directly comparable across tokenizers with different vocabularies and token lengths (see Appendix~\ref{app:metrics}).

\section{Results}
\subsection{Qualitative Analysis}
 Here, we summarize the main observations and representative examples regarding our qualitative analysis on (i)  how each tokenizer segments sentences containing common multiword expressions and (ii)  the sets of frequent tokens that are exclusive to each tokenizer across the evaluation datasets.
 Segmentation examples are reported in Appendix~\ref{app:examples} and the   exclusive-token tables in Appendix~\ref{app:exclusive-tokens}.

\begin{table}[t!]
\centering 
\scalebox{0.76}{
\setlength{\tabcolsep}{3pt}
\begin{tabular}{lrrr}
\toprule
\textbf{Dataset} & \textbf{BPE} & \textbf{SuperBPE} & \textbf{\myalgo} \\
\midrule
CSQA   & 16.63  & 12.43 (25.3\%)           & \textbf{12.24 (26.4\%)} \\
latex-formulas  & 68.58  & 59.49 (13.3\%)           & \textbf{54.88 (20.0\%)} \\
SQuAD           & 163.63 & \textbf{125.11 (23.5\%)} & 126.03 (23.0\%) \\
DROP            & 264.26 & \textbf{206.78 (21.7\%)} & 210.43 (20.4\%) \\
Image2Struct    & 247.32 & 212.43 (14.1\%)          & \textbf{207.96 (15.9\%)} \\
\midrule 
\textit{avg decr. w.r.t. BPE} & --  & \textit{19.58}\% & \textit{21.14}\% \\
\bottomrule
\end{tabular}
}
\caption{Average Encoding Length (AEL) comparison between BPE, SuperBPE, and \myalgo. 
Numbers in parentheses indicate the relative reduction with respect to the BPE baseline. Best results are bolded.\vspace{-4mm}}
\label{tab:ael_results}
\end{table}

Two consistent patterns emerge across the   analysis.
First, \myalgo preserves \LaTeX{} command structures more effectively than SuperBPE.  
For instance, the expressions  \texttt{\textbackslash frac} and  \texttt{\textbackslash begin\{equation\}}  
are split by SuperBPE into two tokens while \myalgo represents them as single tokens, and so it does even with more complex expressions such as  
\texttt{\textbackslash sum\_\{i=}.  
Our   exclusive-token analysis (results reported in Appendix \ref{app:exclusive-tokens}, Tables \ref{tab:exclusive_latex}--\ref{tab:exclusive_img2struct})  confirms this at the vocabulary level: \myalgo's most frequent tokens on latex-formulas and Image2Struct include full command structures such as \texttt{\textbackslash begin\{align*\}}, \texttt{\textbackslash end\{align*\}}, \texttt{\textbackslash begin\{equation*\}}, whereas SuperBPE's exclusive tokens on the same datasets are dominated by short, isolated symbol pairs (\texttt{left(}, \texttt{\textbackslash right}, \texttt{(t}, \texttt{(z}).
Second, on natural-language text, 
\myalgo reveals to be much more effective than SuperBPE  in  preserving multiword expressions as single, indivisible units, 
retaining the semantic and functional integrity of these expressions. 
Surprisingly, this holds not only for relatively frequent and highly lexicalized terms, like  \texttt{"state of the art"} or \texttt{"artificial intelligence"}, but also for rarer, more compositional-looking or 
extended prepositional phrases, such as  \texttt{"late at night"}, \texttt{"in the ground"}, \texttt{"at the end of"}, \texttt{"on the side of the road"}, 
which are not captured by SuperBPE. 
The exclusive-token results (Appendix \ref{app:exclusive-tokens}) show that this is a systematic, dataset-wide phenomenon. 
 Thus, \myalgo improves upon SuperBPE in  avoiding the loss of idiomatic or collocational meaning that arises from purely compositional tokenization,  yielding more efficient, semantically coherent input units, with potential advantages  for downstream language understanding and generation tasks (cf. Section \ref{sec:downstream}).

\subsection{Average Encoding Length}

As shown in Table~\ref{tab:ael_results}, \myalgo consistently achieves the lowest AEL on latex-domain datasets (i.e., latex-formulas and Image2Struct), with a relative length reduction up to $7\%$ vs.  SuperBPE. Also, on natural language benchmarks, \myalgo performs competitively with SuperBPE, showing that it successfully retains the compression advantages of the SuperBPE baseline while offering superior specialization for mathematical notation.

\begin{table}[t!]
\centering
\small
\scalebox{0.9}{
\begin{tabular}{lrrr}
\toprule
\textbf{Dataset} & \textbf{BPE} &
\textbf{SuperBPE} &
\textbf{\myalgo} \\
\midrule
CSQA & 1.81 & 1.68 & 1.77 \\
latex-formulas & 34.22 & 25.52 & 25.49 \\
SQuAD & 74.41 & 68.54 & 68.77\\
DROP & 10.88  & 8.27  & 8.32 \\
Image2Struct & 0.63 & 0.44 & 0.42 \\
\bottomrule
\end{tabular}
}
\caption{Average tokenization time (in seconds).
\vspace{-3mm}}
\label{tab:tokenization-time}
\end{table}

\vspace{1mm}
\noindent\textbf{Time Efficiency.}
Table~\ref{tab:tokenization-time} reports the average tokenization time per instance for the   tokenizers across   datasets.\footnote{Measured on a machine with Ubuntu, Intel(R) Xeon(R) CPU @ 2.20GHz, 30GB RAM.  
\myalgo is implemented in Python using the \textit{PySpark} library. 
} 
  SuperBPE and \myalgo consistently achieve lower tokenization times than   BPE, indicating that the learned supertokens reduce the number of tokenization operations required at inference time. The largest absolute improvements are   on the longest and most token-intensive datasets, i.e., latex-formulas and SQuAD, where tokenization time is reduced by roughly 25\% and 8\%, respectively. 
The differences between SuperBPE and \myalgo are negligible overall, with \myalgo exhibiting a slight advantage on the technical datasets and SuperBPE performing marginally better on the general-language datasets. This comparable efficiency confirms that \textit{the \myalgo's dynamic-frequency-aware tokenization approach affects the learned vocabulary but substantially not  the computational complexity of the tokenization algorithm.}

\vspace{-1mm}
\subsection{Supertoken contribution}

\begin{table}[t!]
\centering
\small
\scalebox{0.9}{
\setlength{\tabcolsep}{3pt}
\begin{tabular}{llccc}
\toprule
& \textbf{Tokenizer} & \textbf{Super Share} & \textbf{Mean Savings} & \textbf{Total Savings} \\
\midrule
\multirow{3}{*}{\rotatebox[origin=c]{90}{\textit{\textbf{latex}}}}
& SuperBPE & 0.133 & 1.169 & 0.156 \\
& \myalgo & \textbf{0.272} & \textbf{2.094} & \textbf{0.569} \\
& $\Delta$ (\%) & +104.5\% & +79.1\% & +264.7\% \\
\midrule
\multirow{3}{*}{\rotatebox[origin=c]{90}{\textit{\textbf{I2S}}}}
& SuperBPE & 0.147 & 1.134 & 0.167 \\
& \myalgo & \textbf{0.366} & \textbf{2.049} & \textbf{0.750} \\
& $\Delta$ (\%) & +149.0\% & +80.7\% & +349.1\% \\
\midrule
\multirow{3}{*}{\rotatebox[origin=c]{90}{\textit{\textbf{DROP}}}}
& SuperBPE & 0.229 & 1.226 & 0.281 \\
& \myalgo & \textbf{0.715} & \textbf{2.836} & \textbf{2.029} \\
& $\Delta$ (\%) & +212.2\% & +131.3\% & +622.1\% \\
\midrule
\multirow{3}{*}{\rotatebox[origin=c]{90}{\textit{\textbf{SQuAD}}}}
& SuperBPE & 0.257 & 1.209 & 0.311 \\
& \myalgo & \textbf{0.822} & \textbf{2.947} & \textbf{2.424} \\
& $\Delta$ (\%) & +219.8\% & +143.8\% & +679.4\% \\
\midrule
\multirow{3}{*}{\rotatebox[origin=c]{90}{\textit{\textbf{CSQA}}}}
& SuperBPE & 0.281 & 1.206 & 0.339 \\
& \myalgo & \textbf{0.836} & \textbf{3.150} & \textbf{2.634} \\
& $\Delta$ (\%) & +197.5\% & +161.2\% & +677.0\% \\
\bottomrule
\end{tabular}
}
\vspace{-1mm}
\caption{Comparison of \myalgo and SuperBPE in terms of  Supertoken Share, Mean Savings, and Total Savings.   
Best results per dataset are in bold.\vspace{-5mm}}
\label{tab:tokenizer_performance}
\end{table}

Table~\ref{tab:tokenizer_performance} compares the token savings achieved by SuperBPE and \myalgo across the evaluation datasets. 
\myalgo consistently outperforms SuperBPE on all three criteria, demonstrating the effectiveness of our proposed dynamic-frequency-aware tokenization  strategy. 
In particular, \myalgo substantially increases the share of text represented by supertokens, with improvements ranging from about 105\% on \textit{latex-formulas} to over 200\% on DROP and SQuAD. 
This larger supertoken utilization translates into considerably higher average and total token savings. The improvements are especially pronounced for the most token-intensive datasets, where \myalgo achieves up to 161\% higher mean savings and over 670\% higher total savings compared with SuperBPE. 

These results indicate that \textit{\myalgo is able to identify and exploit more effective multi-token patterns, leading to a more compact representation of the input text.} 
Notably, the gains are consistent across both technical datasets (e.g., \textit{latex-formulas} and Image2Struct) and general-language benchmarks (e.g., DROP, SQuAD, and CSQA), suggesting that \myalgo provides a general improvement over existing supertokenization approach.

\vspace{-1mm}
\subsection{Token distribution analysis} 
Table~\ref{tab:ngramd} shows that   \myalgo  achieves significantly higher lexical diversity than the BPE baseline, while remaining  comparable to SuperBPE across all datasets. 
Moreover, both SuperBPE and \myalgo improve over BPE in distributional balance, with SuperBPE retaining a small but consistent edge over \myalgo. 
These results are in relation with those from the supertoken analysis of Table~\ref{tab:tokenizer_performance}: \myalgo exhibits a much higher supertoken share on these datasets, meaning it encodes a larger portion of a text with frequently recurring supertokens. Because these units are, by construction, high-frequency patterns, they recur more often in the token stream, which marginally lowers the proportion of unique n-grams. In other words, \myalgo trades a small amount of n-gram diversity 
and token-frequency balance for a substantially higher compression of recurrent expressions.

\begin{table}[t!]
\centering
\scalebox{0.75}{
\setlength{\tabcolsep}{3pt}
\begin{tabular}{lccccccccc}
\toprule
\textbf{Dataset} & 
\multicolumn{3}{c}{\textbf{BPE}}
&
\multicolumn{3}{c}{\textbf{SuperBPE}}
&
\multicolumn{3}{c}{\textbf{\myalgo}} \\
& NDiv & RE$_1$ &  RE$_2$ 
& NDiv & RE$_1$ &  RE$_2$
& NDiv & RE$_1$ &  RE$_2$
\\
\midrule
CSQA & 2.05 & 0.70 & 0.50 & 2.72 &  0.80 & 0.51 & 2.67 & 0.80 & 0.50  \\
latex & 0.48 & 0.59 & 0.48 & 0.72 & 0.63 & 0.52 & 0.74 & 0.65 & 0.52 \\
SQuAD & 0.49 & 0.66 & 0.41 & 0.67 & 0.80 & 0.57 & 0.66 & 0.80 & 0.56 \\
DROP & 0.17 & 0.67 & 0.45 & 0.23 & 0.78 & 0.59 & 0.22 & 0.77 & 0.57  \\
I2S & 0.98 & 0.67 & 0.51 & 1.26 & 0.72 & 0.57 & 1.23 & 0.72 & 0.53  \\
\bottomrule
\end{tabular}
}
\vspace{-1mm}
\caption{Comparison over N-gram diversity (NDiv) and Rényi efficiency (RE$_\alpha$, with $\alpha$ = \{\!1,2\}).\vspace{-3mm}} 
\label{tab:ngramd}
\end{table}

\section{Sensitivity Analysis}
\vspace{-1mm}

We study the sensitivity of \myalgo to the two language-dependent parameters $\theta_{\mathtt{general}}^{\mathtt{min}}$ and $\theta_{\mathtt{latex}}^{\mathtt{min}}$, as shown in Table~\ref{tab:ael_settings} in Appendix~\ref{app:sensitivity}.

Several observations follow. First, the parameters trade off vocabulary size against compression: as the thresholds decrease, more supertokens are added and the AEL drops, but the vocabulary grows rapidly --- from $168$k tokens in the most restrictive setting to more than $1.6$M tokens in the most permissive one. Second, the AEL on natural-language datasets (DROP, SQuAD, CSQA) is governed almost exclusively by $\theta_{\mathtt{general}}^{\mathtt{min}}$: rows with the same value of $\theta_{\mathtt{general}}^{\mathtt{min}}$ but different values of $\theta_{\mathtt{latex}}^{\mathtt{min}}$ yield identical AEL on these three datasets. Symmetrically, $\theta_{\mathtt{latex}}^{\mathtt{min}}$ controls the AEL on \LaTeX-heavy datasets (latex-formulas, Image2Struct). This confirms the design intent of the algorithm, which is to decouple the two language regimes. Finally, very aggressive settings (last row) achieve large AEL reductions but require vocabularies an order of magnitude larger than the baseline, which is unlikely to be practical in real training scenarios.

\vspace{-1mm}
\section{Downstream validation}
\vspace{-1mm}
\label{sec:downstream}
Table \ref{tab:downstream} summarizes our main results on the impact of \myalgo and SuperBPE on downstream performance. 
Looking at the general-language benchmarks (i.e., BoolQ, CSQA and Winogrande), \textit{OLMo2-\myalgo always achieves higher benchmark-accuracy than OLMo2-SuperBPE}, while incurring a slight BPB penalty on two  of the three datasets. 
Results indicate that \myalgo's tokenization does not come at the cost of downstream task performance on general natural language.  
On GSM8K and MATH500, both models exhibit a floor effect in accuracy due to their limited scale, but \myalgo achieves a \textit{clear BPB advantage}.

Note that downstream performance was evaluated with greedy decoding, so results are not influenced by stochastic sampling at inference time. 
It  should also be emphasized that, for OLMo2-\myalgo, we adopt the  \myalgo configuration that was selected for its encoding efficiency as found in our sensitivity analysis (i.e., $\theta_{\text{latex}}^{\min}=0.05$, $\theta_{\text{general}}^{\min}=0.25$, along with $\mathtt{min\_thr} = 10^{-4}$ and  $I=3$); however, this setting is not necessarily optimal for downstream performance. 
Nonetheless, the presented downstream validation analysis has  unveiled that the encoding efficiency advantages brought from \myalgo translate to comparable or improved downstream behavior w.r.t.     SuperBPE.

\begin{table}[t!]
\centering
\small
\setlength{\tabcolsep}{5pt}
\scalebox{0.9}{ 
\begin{tabular}{lcccc}
\toprule
\multirow{2}{*}{\textbf{Dataset}} & \multicolumn{2}{c}{\textbf{OLMo2-\myalgo}} & \multicolumn{2}{c}{\textbf{OLMo2-SuperBPE}} \\ 
& Acc. & BPB & Acc.  & BPB \\ 
\midrule
GSM8K  & na  & \textbf{1.51} & na & 1.56 \\ 
MATH500  & na  & \textbf{1.21} & na & 1.27 \\ 
BoolQ  & \textbf{0.6171}  & 1.84 & 0.6150 & \textbf{1.44} \\ 
CSQA  & \textbf{0.1966} & 8.27 & 0.1949 &  \textbf{7.30} \\ 
Winogrande  & \textbf{0.5043}  & \textbf{0.56} & 0.4972 & 0.58 \\ \bottomrule
\end{tabular}
}
\vspace{-1mm}
\caption{Downstream task results.\vspace{-3mm}}
\label{tab:downstream}
\end{table}

\section{Conclusions} 
\vspace{-1mm}
We present \myalgo, an adaptive tokenization method that extends BPE through frequent-pattern mining to identify multiword expressions and construct an enhanced vocabulary. 
\myalgo{} dynamically adjusts token merge criteria based on both pattern length and type, explicitly distinguishing general-language patterns from formal-language structures, with a focus on \LaTeX{} expressions. 
Our qualitative and quantitative evaluations, including comparison with SuperBPE, show that \myalgo{} more effectively captures both \LaTeX{} commands and general-language multiword expressions as coherent units, preserving their semantic and functional integrity. It achieves up to 7\% relative encoding reduction on mathematical \LaTeX{} content while maintaining competitive natural-language performance. Experiments with small decoder models further demonstrate improved accuracy and lower bits-per-byte on mathematical benchmarks, with comparable results on general-language tasks.

We aim  to investigate the extension of \myalgo beyond \LaTeX{} to other formal (markup) languages, task-oriented hyperparameter optimization to improve downstream performance,  and scalability of \myalgo  on larger models.

\clearpage

\appendix

\section*{Appendix A: Evaluation Datasets}
\label{app:datasets}

We provide a brief description of the datasets used for the  evaluation of \myalgo and competing tokenizers, along with boxes illustrating example instances. 
Table \ref{tab:dataset_stats} summarizes main statistics of the datasets.

\vspace{1mm}
\noindent\textbf{DROP} \cite{dua-etal-2019-drop} contains questions over passages that potentially require discrete reasoning over multiple pieces of information in the passage.

\begin{tcolorbox}[
    colback=gray!10, colframe=black,
    center title, title=\textit{DROP Example Instance},
    breakable, 
    fonttitle=\bfseries
]
{\small
\textbf{Passage}: To start the season, the Lions traveled south to Tampa, Florida to take on the Tampa Bay Buccaneers. The Lions scored first in the first quarter with a 23-yard field goal by Jason Hanson. The Buccaneers tied it up with a 38-yard field goal by Connor Barth, then took the lead when Aqib Talib intercepted a pass from Matthew Stafford and ran it in 28 yards. The Lions responded with a 28-yard field goal. In the second quarter, Detroit took the lead with a 36-yard touchdown catch by Calvin Johnson, and later added more points when Tony Scheffler caught an 11-yard TD pass. 
\textit{(Cont.)}

\vspace{0.2cm}
\textbf{Question}: How many points did the Buccaneers need to tie in the first?

\vspace{0.2cm}
\textbf{Answer spans}:
\begin{verbatim}
{
  "spans": ["3"],
  "types": ["number"]
}
\end{verbatim}
}
\end{tcolorbox}

\vspace{1mm}
\noindent\textbf{latex-formulas} \cite{latex-formulas-dataset} consists of mathematical expressions written in \LaTeX{}.

\begin{tcolorbox}[
    colback=gray!10, colframe=black,
    center title, title=\textit{latex-formulas Example Instance},
    breakable, fonttitle=\bfseries
]
{\small
\begin{verbatim}
\begin{align*}
a^{\dagger}_m=\sqrt{\frac{2}{\zeta}}z_m\,,
\quad a_m=\sqrt{\frac{2}{\zeta}}\bar{z}_m
\quad\textrm{for}\quad m=1,\,3\,,
\end{align*}
\end{verbatim}
}
\end{tcolorbox}

\vspace{1mm}
\noindent\textbf{SQuAD} \cite{rajpurkar-etal-2016-squad} contains passages paired with questions about their content; the answer is always a contiguous span of the corresponding passage.

\begin{tcolorbox}[
    colback=gray!10, colframe=black,
    center title, title=\textit{SQuAD Example Instance},
    breakable, fonttitle=\bfseries
]
{\small

\textbf{Context}: Architecturally, the school has a Catholic character. Atop the Main Building's gold dome is a golden statue of the Virgin Mary. Immediately in front of the Main Building and facing it, is a copper statue of Christ with arms upraised with the legend ``Venite Ad Me Omnes''. Next to the Main Building is the Basilica of the Sacred Heart. Immediately behind the basilica is the Grotto, a Marian place of prayer and reflection. 
\textit{(Cont.)}

\vspace{0.2cm}
\textbf{Question}: What sits on top of the Main Building at Notre Dame?

\vspace{0.2cm}
\textbf{Gold answer}: ``a golden statue of the Virgin Mary'' \\
\textbf{answer\_start}: 92
}
\end{tcolorbox}

\vspace{1mm}
\noindent\textbf{Image2Struct} \cite{roberts2024image2struct} contains \LaTeX{} code extracted from arXiv papers. For our experiments, we use the \textit{equation} and \textit{algorithm} subsets.

\begin{tcolorbox}[
    colback=gray!10, colframe=black,
    center title, title=\textit{Image2Struct Example Instance},
    breakable, fonttitle=\bfseries
]
{\small

\textbf{Text}:
\begin{verbatim}
\begin{algorithmic}
\State $PA\_set$
\For { OC\_block in OC\_blocks}
\State $PA\_block$
\For {$B_i \in$ OC\_block}
...
(Cont.)
...
\mathcal{F}_{a}^{PA}$
\EndFor
\State $PA\_set \gets PA\_block$
\EndFor
\end{algorithmic}
\end{verbatim}
}
\end{tcolorbox}

\vspace{1mm}
\noindent\textbf{CommonsenseQA} (CSQA) \cite{talmor-etal-2019-commonsenseqa} contains 5-way multiple-choice questions that require commonsense knowledge to answer.

\begin{tcolorbox}[
    colback=gray!10, colframe=black,
    center title, title=\textit{CSQA Example Instance},
    breakable, fonttitle=\bfseries
]
{\small

\textbf{Question}: The sanctions against the school were a punishing blow, and they seemed to what the efforts the school had made to change?

\vspace{0.2cm}
\textbf{Choices}:
\begin{verbatim}
A: ignore
B: enforce
C: authoritarian
D: yell at
E: avoid
\end{verbatim}

\vspace{0.2cm}
\textbf{Answer}: A
}
\end{tcolorbox}

\begin{table}[h!]
\centering
\renewcommand{\arraystretch}{0.75} 
\setlength{\tabcolsep}{1.5pt} 
\scalebox{0.74}{
\begin{tabular}{lrrrrr}
\toprule
\textbf{Statistic} & \textbf{DROP} & \textbf{Image2Struct} & \textbf{CSQA} & \textbf{latex-forms.} & \textbf{SQuAD} \\
\midrule
\# Instances & 9,535 & 600 & 9,741 & 100,000$^{(*)}$  & 87,599 \\
\midrule
\multicolumn{6}{l}{\textit{Characters per instance}} \\
Min & 290 & 39 & 20 & 29 & 179 \\
Max & 5,578 & 3,869 & 405 & 440 & 26,203 \\
Avg & 1,111.27 & 632.50 & 78.84 & 145.05 & 814.94 \\
Std & 598.84 & 634.03 & 30.43 & 74.97 & 320.16 \\
\midrule
\multicolumn{6}{l}{\textit{BPE tokens per instance}} \\
Min & 75 & 14 & 5 & 10 & 31 \\
Max & 1,290 & 1,720 & 73 & 294 & 913 \\
Avg & 264.26 & 247.32 & 16.63 & 68.58 & 163.63 \\
Std & 125.25 & 238.77 & 6.72 & 36.58 & 63.14 \\
\midrule
\multicolumn{6}{l}{\textit{Words per instance}$^{(**)}$} \\
Min & 46 & 2 & 4 & 1 & 25 \\
Max & 957 & 536 & 66 & 136 & 666 \\
Avg & 186.38 & 64.42 & 14.60 & 9.72 & 129.82 \\
Std & 100.22 & 68.10 & 5.61 & 8.76 & 49.55 \\
\midrule
\multicolumn{6}{l}{\textit{Alphanumeric BPE tokens per instance}$^{(***)}$} \\
Min & 3 & 0 & 0 & 0 & 0 \\
Max & 90 & 145 & 3 & 94 & 97 \\
Avg & 25.74 & 11.11 & 0.01 & 5.18 & 5.81 \\
Std & 14.86 & 15.61 & 0.11 & 6.04 & 7.07 \\
\midrule
\multicolumn{6}{l}{\textit{\LaTeX{} BPE tokens per instance}} \\
Min & --- & 14 & --- & 10 & --- \\
Max & --- & 1,720 & --- & 294 & --- \\
Avg & --- & 247.32 & --- & 68.58 & --- \\
Std & --- & 238.77 & --- & 36.58 & --- \\
\bottomrule
\end{tabular}
}
\caption{Summary statistics of the evaluation  datasets. 
$^{(*)}$ The full dataset contains approximately one million instances; for evaluation we selected the first $100{,}000$.
$^{(**)}$ Words are regarded as sequences of characters separated by whitespace. $^{(***)}$ BPE tokens containing at least one digit.
}
\label{tab:dataset_stats}
\end{table}

\section*{Appendix B: Assessment Criteria}
\refstepcounter{section}
\label{app:metrics}

\vspace{1mm}
\noindent\textbf{Average Encoding Length.}
Given a corpus $\mathcal{C}$ and a tokenizer $T$, the \textit{Average Encoding Length} (AEL)  is the average number of tokens produced by   $T$ on instances of $\mathcal{C}$:
\begin{equation}
    AEL(\mathcal{C}, T)= \frac{1}{|\mathcal{C}|}\sum_{s \in \mathcal{C}} |T(s)|,
\label{eq:ael}
\end{equation}
  where $T(s)$ denotes the set of tokens of text $s$.

\vspace{1mm}
\noindent\textbf{Supertoken contribution criteria.} 
Let $V$ be the vocabulary created by  a tokenizer (i.e.,  \myalgo or SuperBPE, in our setting)  built on top of an existing   BPE vocabulary. 
We can partition $V$ into the set of \emph{BPE tokens} $V_{\text{BPE}}$, which are the tokens inherited from the underlying BPE vocabulary, and the set of  \emph{supertokens} $V_{\text{super}}$. 
For any token $t \in V$  and a corpus $\mathcal{C}$, let $\textit{count}(t,T(\mathcal{C}))$ denote the number of occurrences of $t$ in the encoded corpus $T(\mathcal{C})$,  and $\textit{depth}(t)$ the number of base BPE tokens  constituting   $t$.   
Note that any token $t$ is a supertoken iff $\textit{depth}(t) > 1$;  also,   a supertoken may itself be formed by merging two existing supertokens---as happens  at later iterations of \myalgo---but this does not affect the definition, which depends only on the full decomposition of $t$ into BPE tokens.

We define three metrics to evaluate the behavior of supertokens:
\begin{enumerate}
    \item \textbf{Supertoken Share} ($\text{Share}_{\text{super}}$): the fraction of the encoded token stream that consists of supertokens,
    \begin{equation}
    \small
        \text{Share}_{\text{super}}(\mathcal{C}, T) = \frac{\sum_{t \in V_{\text{super}}} \textit{count}(t,T(\mathcal{C}))}{\sum_{t \in V} \textit{count}(t,T(\mathcal{C}))}.
    \label{eq:share}
    \end{equation}
    
    \item \textbf{Mean Savings} ($\mu_{\text{savings}}$):  the expected number of base BPE tokens saved per supertoken occurrence, 
    \begin{equation}
    \small
    \hspace{-8mm}    \mu_{\text{savings}}(\mathcal{C}, T) = \frac{\sum_{t \in V_{\text{super}}} \textit{count}(t,T(\mathcal{C})) \cdot (\textit{depth}(t) - 1)}{\sum_{t \in V_{\text{super}}} \textit{count}(t,T(\mathcal{C}))}.
    \end{equation}
    
    \item \textbf{Total Savings} ($\text{Savings}_{\text{total}}$): the average number of BPE tokens saved per emitted token over the entire corpus, which quantifies the overall compression gain,
    \begin{equation}
    \small
        \text{Savings}_{\text{total}}(\mathcal{C}, T) = \text{Share}_{\text{super}}(\mathcal{C}, T) \cdot \mu_{\text{savings}}(\mathcal{C}, T).
        \label{eq:totalsavings}
    \end{equation}
\end{enumerate}

\vspace{1mm}
\noindent\textbf{N-gram Diversity} 
Given a corpus $\mathcal{C}$, a tokenizer $T$, and a maximum n-gram order $N$, the N-gram diversity is defined as the sum, over all n-gram orders from 1 to $N$, of the fraction of distinct n-grams among all n-grams extracted from the tokenized corpus:

\begin{equation}
\small
    NgD(\mathcal{C}, T, N)=\sum_{n=1}^N\frac{\#\text{unique-$n$-grams} \in T(\mathcal{C})}{\#\text{$n$-grams} \in T(\mathcal{C})}    
\label{eq:ngd}
\end{equation}

\noindent
In our experiments, $N$ was set to 4.

\vspace{1mm}
\noindent\textbf{Rényi Efficiency}
Let $p(t,T, \mathcal{C}) = \frac{\text{count}(t,T(\mathcal{C}))}{|T(\mathcal{C})|}$ be the relative frequency of token $t$ in the tokenized corpus. The Rényi efficiency is defined as:
\begin{equation}
\small
    RE(\mathcal{C}, T, \alpha) = \lim_{\alpha' \to \alpha} \frac{1}{(1-\alpha')} \ln \left( \sum_{t} p(t,T, \mathcal{C})^{\alpha'} \right),
\label{eq:renyi}
\end{equation}

\noindent 
where  $\alpha>0$ controls the sensitivity  to the tail of the token-frequency distribution, and how strongly frequent tokens dominate the score. 
In particular: 
\begin{itemize}
    \item $\alpha = 1$:  Taking the limit to 1 recovers the standard Shannon-entropy-based measure, $ -\sum_{t} p(t,T, \mathcal{C}) \ln p(t,T, \mathcal{C})$
    \item $\alpha = 2$:  This  gives more weight to the most frequent tokens via squared probabilities.
\end{itemize}

\vspace{1mm}
\noindent\textbf{Bits-per-byte (BPB)} Given a tokenizer $T$ and an input text sequence $s$, Bits-per-byte (BPB)  
is a standard criterion  for evaluating language models and compressors, which is defined as:

\begin{equation}
\small
    BPB(s,T)= \frac{\mathcal{L}_{CE}(T(s))}{\ln(2) \cdot n_{\texttt{bytes}}(T(s))},
\end{equation}
where  $T(s)$ is the set of tokens  obtained by tokenizing the input text $s$, $n_{\texttt{bytes}}(T(s))$ is the number of bytes in the original text represented by the token sequence, and 
$\mathcal{L}_{CE}$ is the cross-entropy loss of $T(s)$.

\section*{Appendix C: Training Data for the Tokenizers}
\refstepcounter{section}
\label{app:training}
All tokenizers were trained on a 10GB subset of the OLMo2 stage-2 pretraining
corpus (\emph{olmo-mix}) \citep{OLMo20242O2} publicly released by
\citet{liu-etal-2025-superbpe}. This 10GB subset is drawn from the full released
data, which totals approximately 100GB; we used the smaller 10GB portion for
tokenizer training as-is, without further filtering. As reported by
\citet{liu-etal-2025-superbpe}, a small amount of tokenizer training data is
sufficient---enlarging it beyond this scale does not measurably affect the
encoding efficiency of the resulting tokenizer---and the subset already
incorporates their document-truncation preprocessing, which limits the influence
of a few extremely long, internally duplicated documents. Using the same released
data for BPE, SuperBPE, and \myalgo ensures that any difference in the resulting
tokenizers is due to the tokenization algorithm rather than from the training
data, and makes our setup directly reproducible.

To determine \texttt{min\_thr}, $\theta^{\min}_{\text{general}}$, and
$\theta^{\min}_{\text{latex}}$, we estimated the bigram frequency distributions of Figures~\ref{fig:minthr} and~\ref{fig:splitdist} from a random sample of this 10GB corpus. Sampling was performed at the file level: files were drawn uniformly at random from the 10GB subset, and processed in this random order until the cumulative number of extracted bigrams reached $1.5 \cdot 10^8$.

\vspace{1mm}
\noindent\textbf{Remarks on the selection of hyperparameters.} 
The hyperparameters of our \myalgo  were determined through a data-driven procedure using a sample drawn from the same training distribution employed for SuperBPE, rather than being manually tuned for a specific downstream setting. This design choice ensures that the selected frequency thresholds reflect the statistical properties of the data while avoiding overfitting to the full training corpus. Since the procedure relies on identifying recurring patterns and adapting the thresholds accordingly, rather than on corpus-specific heuristics, we expect the resulting methodology to generalize to other datasets with different distributions and domains. In this sense, the proposed approach provides a robust and transferable strategy for configuring pattern-aware BPE variants.

\section*{Appendix D: Qualitative Comparison}
\refstepcounter{section}
\label{app:examples}

To illustrate the segmentation behavior of \myalgo{} on representative multiword expressions, we generated  a set of example sentences targeting constructs known to be challenging for whitespace-based subword tokenization. The generated sentences were then encoded using the actual trained  SuperBPE and \myalgo{} tokenizers, along with basic BPE. 

The resulting token sequences are reported verbatim in the following listings, for each of the tokenizers.  
Underlined terms correspond to compound words or named entities that are expected to be encoded as single tokens.

\definecolor{bpeblue}{HTML}{E3F2FD}
\definecolor{semgreen}{HTML}{E8F5E9}
\definecolor{freqred}{HTML}{F88379}
\newcommand{\bpetoken}[1]{\colorbox{bpeblue}{\small\textsf{#1}}}
\newcommand{\supertoken}[1]{\colorbox{semgreen}{ {\small\textsf{#1} } }}
\newcommand{\freqtoken}[1]{\colorbox{freqred}{ {\small \textsf{#1} }}}

{
\small
\begin{center}

\vspace{0.5cm}
\textbf{\textsc{Example 1 ---} \textbackslash frac\{n \textbackslash sum\_\{i=1\}\textasciicircum n i\}\{N\} \textbackslash rightarrow x}
\vspace{0.3cm}

\begin{tabular}{@{}>{\bfseries}l@{\hspace{1em}}p{0.78\linewidth}@{}}
\toprule
BPE & 
\bpetoken{\textbackslash} 
\bpetoken{frac} \bpetoken{\{}  \bpetoken{n}  
\bpetoken{\textbackslash} 
\bpetoken{sum} 
\bpetoken{\_\{} 
\bpetoken{i} 
\bpetoken{=} 
\bpetoken{1}
\bpetoken{\}\textasciicircum}  \bpetoken{n} 
\bpetoken{Ġi} 
\bpetoken{\}\{\textbackslash}  \bpetoken{N} 
\bpetoken{\}} 
\bpetoken{Ġ\textbackslash} 
\bpetoken{rightarrow} 
\bpetoken{Ġx}  
\\
\addlinespace
SuperBPE & 
\supertoken{\textbackslash} \supertoken{frac\{} 
\supertoken{n} 
\supertoken{\textbackslash} 
\supertoken{sum}  
\supertoken{\_\{i} 
\supertoken{=} 
\supertoken{1}
\supertoken{\}\textasciicircum}  \supertoken{n} 
\supertoken{Ġi} 
\supertoken{\}\{\textbackslash}  \supertoken{N} 
\supertoken{\}Ġ\textbackslash} 
\supertoken{rightarrow} 
\supertoken{Ġx} 
\\
\addlinespace
AdaBPE & \freqtoken{\textbackslash frac\{} 
\freqtoken{n}
\freqtoken{\textbackslash sum\_\{i=} 
\freqtoken{1}
\freqtoken{\}\textasciicircum}  \freqtoken{n} 
\freqtoken{Ġi} 
\freqtoken{\}\{\textbackslash}  \freqtoken{N} 
\freqtoken{\}Ġ\textbackslash rightarrow} 
\freqtoken{Ġx}  
\\
\bottomrule
\end{tabular}

\vspace{1cm}





\textbf{\textsc{Example 2 ---} \textbackslash begin\{equation\}y = ax + b \textbackslash end\{equation\}}

\vspace{0.3cm}

\begin{tabular}{@{}>{\bfseries}l@{\hspace{1em}}p{0.78\linewidth}@{}}
\toprule
BPE & 
\bpetoken{\textbackslash} 
\bpetoken{begin}
\bpetoken{\{}
\bpetoken{equation}
\bpetoken{\}}
\bpetoken{Ġy}
\bpetoken{Ġ=}
\bpetoken{Ġax}
\bpetoken{Ġ+}
\bpetoken{Ġb}
\bpetoken{Ġ\textbackslash}
\bpetoken{end}
\bpetoken{\{}
\bpetoken{equation}
\bpetoken{\}}
\\
\addlinespace
superBPE & 
\supertoken{{\textbackslash}begin\{equation} \supertoken{\}} \supertoken{Ġy} \supertoken{Ġ=} \supertoken{Ġax} \supertoken{Ġ+} \supertoken{Ġb} \supertoken{Ġ{\textbackslash}}
\supertoken{end\{}
\supertoken{equation}
\supertoken{\}}
\\
\addlinespace
AdaBPE & 
\freqtoken{{\textbackslash}begin\{equation\}} \freqtoken{Ġy} \freqtoken{Ġ=} \freqtoken{Ġax} \freqtoken{Ġ+} \freqtoken{Ġb} \freqtoken{Ġ{\textbackslash}end\{equation\}}
\\
\bottomrule
\end{tabular}

\vspace{1cm}

\textbf{\textsc{Example 3 ---} The development of \underline{Artificial Intelligence} requires vast amounts of data.}
\vspace{0.3cm}

\begin{tabular}{@{}>{\bfseries}l@{\hspace{1em}}p{0.78\linewidth}@{}}
\toprule
BPE & 
\bpetoken{The} \bpetoken{Ġdevelopment} \bpetoken{Ġof} \bpetoken{ĠArtificial} \bpetoken{ĠIntelligence} \bpetoken{Ġrequires} \bpetoken{Ġvast} \bpetoken{Ġamounts} \bpetoken{Ġof} \bpetoken{Ġdata} \bpetoken{.}
\\
\addlinespace
SuperBPE & 
\supertoken{The} \supertoken{ĠdevelopmentĠof} \supertoken{ĠArtificial} \supertoken{ĠIntelligence} \supertoken{Ġrequires} \supertoken{Ġvast} \supertoken{ĠamountsĠof} \supertoken{Ġdata.}
\\
\addlinespace
AdaBPE & \freqtoken{The} \freqtoken{ĠdevelopmentĠof} \freqtoken{ĠArtificialĠIntelligence} \freqtoken{Ġrequires} \freqtoken{Ġvast} \freqtoken{ĠamountsĠof} \freqtoken{Ġdata.}
\\
\bottomrule
\end{tabular}

\vspace{1cm}

\textbf{\textsc{Example 4 ---} The new software uses \underline{state of the art} \underline{machine learning}.}
\vspace{0.3cm}

\begin{tabular}{@{}>{\bfseries}l@{\hspace{1em}}p{0.78\linewidth}@{}}
\toprule
BPE & 
\bpetoken{The}   \bpetoken{Ġnew} \bpetoken{Ġsoftware} \bpetoken{Ġuses} \bpetoken{Ġstate} \bpetoken{Ġof} \bpetoken{Ġthe} \bpetoken{Ġart} \bpetoken{Ġmachine} \bpetoken{Ġlearning} \bpetoken{.}
\\
\end{tabular}

\begin{tabular}{@{}>{\bfseries}l@{\hspace{1em}}p{0.78\linewidth}@{}}
SuperBPE & 
\supertoken{TheĠnew} \supertoken{Ġsoftware} \supertoken{Ġuses} \supertoken{ĠstateĠofĠthe} \supertoken{Ġart} \supertoken{ĠmachineĠlearning} \supertoken{.}
\\
\addlinespace
AdaBPE & \freqtoken{TheĠnew}  \freqtoken{Ġsoftware} \freqtoken{Ġuses} \freqtoken{ĠstateĠofĠtheĠart} \freqtoken{ĠmachineĠlearning} \freqtoken{.}
\\
\bottomrule
\end{tabular}

\vspace{5mm}

\textbf{\textsc{Example 5 ---} The \underline{high school} student ate \underline{ice cream} while sitting on the \underline{living room} couch.}

\vspace{0.3cm}
\begin{tabular}{@{}>{\bfseries}l@{\hspace{1em}}p{0.78\linewidth}@{}}
\toprule
\addlinespace
BPE & \bpetoken{The} \bpetoken{Ġhigh} 
\bpetoken{Ġschool} \bpetoken{Ġstudent} \bpetoken{Ġate} \bpetoken{Ġice}
\bpetoken{Ġcream} \bpetoken{Ġwhile} \bpetoken{Ġsitting} \bpetoken{Ġon} \bpetoken{Ġthe} \bpetoken{Ġliving} \bpetoken{Ġroom} \bpetoken{Ġcouch} \bpetoken{.} 
\\ 
\addlinespace
SuperBPE & \supertoken{The} 
\supertoken{ĠhighĠschool} 
\supertoken{Ġstudent} 
\supertoken{Ġate} 
\supertoken{ĠiceĠcream} 
\supertoken{Ġwhile}  
\supertoken{ĠsittingĠonĠthe} \supertoken{ĠlivingĠroom} \supertoken{Ġcouch} \supertoken{.} 
\\
\addlinespace
\myalgo & \freqtoken{The} \freqtoken{ĠhighĠschool}  \freqtoken{Ġstudent} \freqtoken{Ġate} \freqtoken{ĠiceĠcream} \freqtoken{Ġwhile} \freqtoken{ĠsittingĠonĠthe} \freqtoken{ĠlivingĠroom} \freqtoken{Ġcouch} 
\\
\bottomrule
\end{tabular}

\vspace{5mm}

\textbf{\textsc{Example 6 ---}  \underline{Storm cloud} gathered over the valley, and another \underline{storm cloud} loomed on the horizon}

\vspace{0.3cm}
\begin{tabular}{@{}>{\bfseries}l@{\hspace{1em}}p{0.78\linewidth}@{}}
\toprule
\addlinespace
BPE & \bpetoken{Storm} 
\bpetoken{Ġcloud} 
\bpetoken{Ġgathered} 
\bpetoken{Ġover} 
\bpetoken{Ġthe} \bpetoken{Ġvalley} \bpetoken{Ġand} \bpetoken{Ġanother} \bpetoken{Ġstorm} \bpetoken{Ġcloud} \bpetoken{Ġloomed} \bpetoken{Ġon} \bpetoken{Ġthe} \bpetoken{Ġhorizon} 
\\
\addlinespace
SuperBPE & \supertoken{Storm} \supertoken{Ġcloud} \supertoken{Ġgathered} \supertoken{ĠoverĠthe} \supertoken{Ġvalley} \supertoken{,ĠandĠanother} \supertoken{Ġstorm} \supertoken{Ġcloud} \supertoken{Ġloomed} \supertoken{ĠonĠthe} \supertoken{Ġhorizon} \\
\addlinespace
\myalgo & \freqtoken{Storm} \freqtoken{Ġcloud} \freqtoken{Ġgathered} \freqtoken{ĠoverĠtheĠvalley} \freqtoken{,ĠandĠanother} \freqtoken{Ġstorm} \freqtoken{Ġcloud} \freqtoken{Ġloomed} \freqtoken{ĠonĠtheĠhorizon} \\
\bottomrule
\end{tabular}

\vspace{5mm}




\textbf{\textsc{Example 7 ---} At the end of the long march, the soldiers were still standing in formation on the side of the road.}

\vspace{0.3cm}
\begin{tabular}{@{}>{\bfseries}l@{\hspace{1em}}p{0.78\linewidth}@{}}
\toprule
\addlinespace
BPE & \bpetoken{At} \bpetoken{Ġthe} \bpetoken{Ġend} \bpetoken{Ġof} \bpetoken{Ġthe} \bpetoken{Ġlong} \bpetoken{Ġmarch} \bpetoken{,} \bpetoken{Ġthe} \bpetoken{Ġsoldiers} \bpetoken{Ġwere} \bpetoken{Ġstill} \bpetoken{Ġstanding}  
\bpetoken{Ġin} \bpetoken{Ġformation} 
\bpetoken{Ġon} \bpetoken{Ġthe} \bpetoken{Ġside} 
\bpetoken{Ġof} 
\bpetoken{Ġthe} \bpetoken{Ġroad} \bpetoken{.} 
\\
\addlinespace
SuperBPE & \supertoken{AtĠthe} 
\supertoken{ĠendĠofĠthe} \supertoken{Ġlong} \supertoken{Ġmarch} \supertoken{,Ġthe} \supertoken{Ġsoldiers}  \supertoken{ĠwereĠstill} \supertoken{Ġstanding} \supertoken{Ġin} \supertoken{Ġformation} \supertoken{ĠonĠthe} \supertoken{ĠsideĠofĠthe} \supertoken{Ġroad} \supertoken{.}
\\
\addlinespace
\myalgo & \freqtoken{AtĠtheĠendĠof} \freqtoken{ĠtheĠlong} \freqtoken{Ġmarch} \freqtoken{,Ġthe} \freqtoken{Ġsoldiers} \freqtoken{ĠwereĠstill} \freqtoken{ĠstandingĠin} \freqtoken{Ġformation}   \freqtoken{ĠonĠtheĠsideĠofĠtheĠroad}  \freqtoken{.}
\\
\bottomrule
\end{tabular}

\vspace{1cm}

\textbf{\textsc{Example 8 ---} Then, late at night, he buried the old box in the ground beside the road.}

\vspace{0.3cm}
\begin{tabular}{@{}>{\bfseries}l@{\hspace{1em}}p{0.78\linewidth}@{}}
\toprule
\addlinespace
BPE & 
\bpetoken{Then} \bpetoken{,} \bpetoken{Ġlate} \bpetoken{Ġat} \bpetoken{Ġnight} \bpetoken{,} \bpetoken{Ġhe} \bpetoken{Ġburied} \bpetoken{Ġthe} \bpetoken{Ġold} \bpetoken{Ġbox} \bpetoken{Ġin} \bpetoken{Ġthe} \bpetoken{Ġground} \bpetoken{Ġbeside} \bpetoken{Ġthe} \bpetoken{Ġroad} \bpetoken{.}
\\
\end{tabular}

\begin{tabular}{@{}>{\bfseries}l@{\hspace{1em}}p{0.78\linewidth}@{}}
SuperBPE & \supertoken{Then}  \supertoken{,} \supertoken{Ġlate} \supertoken{ĠatĠnight} \supertoken{,Ġhe} \supertoken{Ġburied} \supertoken{ĠtheĠold} \supertoken{Ġbox} \supertoken{ĠinĠthe} \supertoken{Ġground} \supertoken{Ġbeside} \supertoken{ĠtheĠroad} \supertoken{.}\\
\addlinespace
\myalgo & \freqtoken{Then} \freqtoken{,} \freqtoken{ĠlateĠatĠnight} \freqtoken{,Ġhe} \freqtoken{Ġburied} \freqtoken{ĠtheĠold} \freqtoken{Ġbox} \freqtoken{ĠinĠtheĠground} \freqtoken{Ġbeside} \freqtoken{ĠtheĠroad} \freqtoken{.} \\
\bottomrule
\end{tabular}

\vspace{1cm}

\end{center}
}

The examples above provide evidence of the recurring differences between \myalgo{} and SuperBPE in their treatment of \LaTeX{} syntax and of general natural language.

For \LaTeX{} examples, \myalgo{} tends to merge complete syntactic units into single tokens, whereas SuperBPE often performs only partial merges. Representative examples include \texttt{\textbackslash begin\{equation\}}, \texttt{\textbackslash end\{equation\}}, \texttt{\textbackslash sum\_\{i=}, and \texttt{\textbackslash rightarrow}.

For general-language examples, the two tokenizers typically agree on short, high-frequency multiword expressions. An exception is Example~3, where \myalgo{} merges \emph{Artificial Intelligence} into a single supertoken, while SuperBPE keeps the two words separate. The difference becomes more pronounced for longer expressions. In Example~4, \myalgo{} merges the entire phrase \emph{state of the art} into a single supertoken, whereas SuperBPE represents it with two tokens. Likewise, in Examples~7--8, \myalgo{} learns longer expressions such as \emph{At the end of}, \emph{on the side of the road}, \emph{late at night}, and \emph{in the ground}, as well as phrasal verbs such as \emph{standing in}, which SuperBPE tokenizes into multiple tokens.

\section*{Appendix E: Exclusive Tokens}
\refstepcounter{section}
\label{app:exclusive-tokens}

To complement our qualitative analysis, we report in Tables~\ref{tab:exclusive_latex}--\ref{tab:exclusive_img2struct}  the top-frequent tokens that are \emph{exclusive} to each tokenizer's vocabulary, on each of the full-size evaluation dataset. 

Two consistent patterns emerge across datasets. On the \LaTeX{}-heavy datasets (latex-formulas, Table~\ref{tab:exclusive_latex};
Image2Struct, Table~\ref{tab:exclusive_img2struct}),   \myalgo{}'s exclusive tokens correspond to well-formed  \LaTeX{} constructs: environment delimiters such as \texttt{begin\{align*\}}, \texttt{end\{align*\}}, indexed-variable patterns such as \texttt{\_\{n}, \texttt{\_\{k}, \texttt{\_\{j}, and command prefixes such as \texttt{\}\textbackslash frac\{} and \texttt{sum\_\{i=} are all represented as single tokens by \myalgo, whereas SuperBPE often splits them into multiple tokens. 

On the general-language datasets (CSQA, Table~\ref{tab:exclusive_commonsense};
SQuAD, Table~\ref{tab:exclusive_squad}; DROP, Table~\ref{tab:exclusive_drop}),  
SuperBPE's unique tokens on these datasets often correspond to dataset-specific named entities or frequent lexicalized expressions, which capture recurring surface forms but do not necessarily reflect productive linguistic patterns (e.g.\ \texttt{UnitedĠStates}, \texttt{AfricanĠAmerican}, \texttt{citationĠneeded}).
\myalgo{}'s exclusive tokens are also  frequent surface forms, but they arguably reflect more compositional and syntactic structure than SuperBPE's dataset-specific named entities, such as determiner-led noun phrases and multiword expressions---e.g.\ \texttt{ĠtheĠregion}, \texttt{ĠtheĠSovietĠUnion}, \texttt{ĠearlyĠin}---consistent with the segmentation examples in Appendix~\ref{app:examples}.

\section*{Appendix F:  Details on Model Training for Downstream Evaluation}

\refstepcounter{section}
\label{app:training_details}

Table~\ref{tab:trainingparameters} reports the main settings used to train OLMo2-\myalgo{} and OLMo2-SuperBPE. 
Both models were trained on the full 100GB subset released by \citet{liu-etal-2025-superbpe}, and 
  share an identical architecture and training budget (e.g., number of epochs, batch size), so that they differ only in the  tokenizer used
to preprocess the training corpus (\myalgo{} vs.\ SuperBPE). 
Keeping every other
architectural and optimization choice fixed isolates the effect of the
tokenizer on the downstream results reported in Table \ref{tab:downstream}  ensuring that
observed differences in accuracy and BPB are attributable to the
tokenization strategy rather than to confounding factors in model capacity
or training regime. 

\begin{table}[h]
    \centering
    \small
    \scalebox{0.9}{
    \begin{tabular}{lcc}
        \toprule
        & \textbf{OLMo-\myalgo} & \textbf{OLMo-SuperBPE} \\
        \midrule
        Architecture & OLMo-2 & OLMo-2 \\
        Hidden size & 2048 & 2048 \\
        Position embedding & RoPE & RoPE \\
        Attention heads & 16 & 16 \\
        Activation function & SiLU & SiLU \\
        Layers & 16 & 16 \\
        Epochs & 3 & 3 \\
        Batch size & 8 & 8 \\
        Optimizer & AdamW & AdamW \\
        Tokenizer & \myalgo & SuperBPE\\
        \bottomrule
    \end{tabular}
    }
    \caption{Main hyperparameters used during model training}
    \label{tab:trainingparameters}
\end{table}

\begin{table*}[t]
\centering
\footnotesize
\setlength{\tabcolsep}{6pt}
\begin{tabular}{rlr@{\hskip 24pt}rlr}
\toprule
\multicolumn{3}{c}{\textbf{\myalgo only}} & \multicolumn{3}{c}{\textbf{SuperBPE only}} \\
\cmidrule(lr){1-3} \cmidrule(lr){4-6}
\textbf{Rank} & \textbf{Token} & \textbf{Frequency} & \textbf{Rank} & \textbf{Token} & \textbf{Frequency} \\
\midrule
1  & \texttt{end\{align*\}}              & 98{,}652 & 1  & \texttt{left(}        & 15{,}092 \\
2  & \texttt{begin\{align*\}}            & 58{,}772 & 2  & \texttt{(t}           & 12{,}354 \\
3  & \texttt{\textbackslash begin\{align}& 40{,}418 & 3  & \texttt{\textbackslash right} & 11{,}718 \\
4  & \texttt{\_\{n}                      & 10{,}119 & 4  & \texttt{,Ġ\textbackslash}     &  9{,}964 \\
5  & \texttt{\_\{k}                      &  6{,}384 & 5  & \texttt{left(\textbackslash}  &  9{,}765 \\
6  & \texttt{\_\{j}                      &  6{,}235 & 6  & \texttt{\textbackslash in}    &  9{,}166 \\
7  & \texttt{Ġ\textbackslash sum\_\{}    &  4{,}486 & 7  & \texttt{n-}           &  8{,}080 \\
8  & \texttt{\}\textbackslash left(}     &  4{,}185 & 8  & \texttt{Ġ+Ġ\textbackslash}    &  7{,}576 \\
9  & \texttt{)Ġ=Ġ\textbackslash}        &  4{,}123 & 9  & \texttt{\}.\textbackslash}    &  7{,}379 \\
10 & \texttt{sum\_\{i=}                  &  4{,}116 & 10 & \texttt{Ġ\textbackslash in}   &  6{,}097 \\
11 & \texttt{tilde\{}                    &  4{,}098 & 11 & \texttt{Ġ\textbackslash right}&  5{,}695 \\
12 & \texttt{int\_}                      &  4{,}077 & 12 & \texttt{(z}           &  5{,}330 \\
13 & \texttt{int\_\{}                    &  4{,}013 & 13 & \texttt{Ġ\&Ġ}         &  5{,}001 \\
14 & \texttt{\}\textbackslash left}      &  3{,}737 & 14 & \texttt{Ġ\&Ġ\textbackslash}   &  4{,}905 \\
15 & \texttt{t\_}                        &  3{,}671 & 15 & \texttt{cases}        &  4{,}492 \\
16 & \texttt{t,}                         &  3{,}449 & 16 & \texttt{i,}           &  3{,}723 \\
17 & \texttt{bar\{}                      &  3{,}436 & 17 & \texttt{(n}           &  3{,}643 \\
18 & \texttt{\}\textbackslash frac\{}    &  3{,}315 & 18 & \texttt{,y}           &  3{,}505 \\
19 & \texttt{\}Ġ-}                       &  3{,}244 & 19 & \texttt{(y}           &  3{,}235 \\
20 & \texttt{Ġx\_}                       &  3{,}190 & 20 & \texttt{(a}           &  2{,}778 \\
\bottomrule
\end{tabular}
\caption{Top exclusive tokens by frequency on {latex-formulas}. Left: tokens used by \myalgo but absent from SuperBPE. Right: tokens used by SuperBPE but absent from \myalgo.}
\label{tab:exclusive_latex}
\end{table*}

\begin{table*}[t]
\centering
\footnotesize
\setlength{\tabcolsep}{6pt}
\begin{tabular}{rlr@{\hskip 24pt}rlr}
\toprule
\multicolumn{3}{c}{\textbf{\myalgo only}} & \multicolumn{3}{c}{\textbf{SuperBPE only}} \\
\cmidrule(lr){1-3} \cmidrule(lr){4-6}
\textbf{Rank} & \textbf{Token} & \textbf{Frequency} & \textbf{Rank} & \textbf{Token} & \textbf{Frequency} \\
\midrule
1  & \texttt{ĠdoĠwhat}      & 201 & 1  & \texttt{WhatĠhappens}      & 80 \\
2  & \texttt{Ġdo?}          & 178 & 2  & \texttt{ĠwasĠhe}           & 67 \\
3  & \texttt{,ĠwhatĠwas}    &  81 & 3  & \texttt{TheĠman}           & 65 \\
4  & \texttt{Ġdidn}         &  72 & 4  & \texttt{WhatĠcan}          & 54 \\
5  & \texttt{ĠwhatĠto}      &  68 & 5  & \texttt{Ġficus}            & 53 \\
6  & \texttt{ĠyouĠbuy}      &  60 & 6  & \texttt{WhyĠwould}         & 51 \\
7  & \texttt{ĠdoingĠwhat}   &  56 & 7  & \texttt{ĠdidĠshe}          & 48 \\
8  & \texttt{Ġdidn'}        &  52 & 8  & \texttt{ĠwereĠthey}        & 43 \\
9  & \texttt{ĠbeĠfound}     &  48 & 9  & \texttt{ĠcouldĠhappen}     & 37 \\
10 & \texttt{WhatĠkindĠof}  &  44 & 10 & \texttt{Ġit.}              & 31 \\
11 & \texttt{ĠstandingĠin}  &  43 & 11 & \texttt{ĠtoĠit}            & 30 \\
12 & \texttt{tĠlike}        &  42 & 12 & \texttt{Ġdidn'tĠwantĠto}   & 22 \\
13 & \texttt{ĠdoĠif}        &  40 & 13 & \texttt{ĠisĠknown}         & 21 \\
14 & \texttt{ĠwhatĠdo}      &  35 & 14 & \texttt{ĠfreshĠair}        & 20 \\
15 & \texttt{ĠdoĠto}        &  34 & 15 & \texttt{ĠwantedĠa}         & 17 \\
16 & \texttt{ĠhappensĠto}   &  32 & 16 & \texttt{WhyĠdoes}          & 15 \\
17 & \texttt{,ĠsoĠwhat}     &  31 & 17 & \texttt{What's}            & 14 \\
18 & \texttt{,ĠwhatĠdo}     &  30 & 18 & \texttt{Ġdidn'tĠhave}      & 13 \\
19 & \texttt{ĠtoĠbed}       &  28 & 19 & \texttt{ĠfastĠfood}        & 13 \\
20 & \texttt{ĠwhatĠwill}    &  25 & 20 & \texttt{Ġdidn'tĠknow}      & 12 \\
\bottomrule
\end{tabular}
\caption{Top exclusive tokens by frequency on {CSQA}.}
\label{tab:exclusive_commonsense}
\end{table*}

\begin{table*}[t]
\centering
\footnotesize
\setlength{\tabcolsep}{6pt}
\begin{tabular}{rlr@{\hskip 24pt}rlr}
\toprule
\multicolumn{3}{c}{\textbf{\myalgo only}} & \multicolumn{3}{c}{\textbf{SuperBPE only}} \\
\cmidrule(lr){1-3} \cmidrule(lr){4-6}
\textbf{Rank} & \textbf{Token} & \textbf{Frequency} & \textbf{Rank} & \textbf{Token} & \textbf{Frequency} \\
\midrule
1  & \texttt{:Ġ}                  & 1{,}735 & 1  & \texttt{citationĠneeded}  & 3{,}488 \\
2  & \texttt{.ĠWhatĠwasĠthe}      & 1{,}664 & 2  & \texttt{?Ġ}               & 2{,}407 \\
3  & \texttt{ĠtheĠlateĠ}          & 1{,}239 & 3  & \texttt{ĠÂ°C}             & 1{,}655 \\
4  & \texttt{:Ġthe}               & 1{,}121 & 4  & \texttt{ĠuntilĠ}          & 1{,}564 \\
5  & \texttt{.ĠWhatĠtypeĠof}      & 1{,}061 & 5  & \texttt{Ġcity's}          & 1{,}214 \\
6  & \texttt{ĠtheĠregion}         & 1{,}044 & 6  & \texttt{ĠapproximatelyĠ}  & 1{,}188 \\
7  & \texttt{Ġ(e}                 &    990 & 7  & \texttt{.ĠByĠ}            &    987 \\
8  & \texttt{ĠtheĠisland}         &    986 & 8  & \texttt{thĠcentury,}      &    981 \\
9  & \texttt{thĠand}              &    973 & 9  & \texttt{ĠMiddleĠAges}     &    981 \\
10 & \texttt{.ĠWhatĠare}          &    922 & 10 & \texttt{th-century}       &    969 \\
11 & \texttt{ĠtheĠperiod}         &    887 & 11 & \texttt{OnĠ}              &    924 \\
12 & \texttt{ĠHowĠmany}           &    874 & 12 & \texttt{thĠcenturies}     &    912 \\
13 & \texttt{ĠtheĠRoman}          &    854 & 13 & \texttt{Ġworld's}         &    880 \\
14 & \texttt{ĠtheĠnational}       &    841 & 14 & \texttt{-speaking}        &    795 \\
15 & \texttt{ĠtheĠSovietĠUnion}   &    833 & 15 & \texttt{ĠUS\$}            &    786 \\
16 & \texttt{ĠtheĠmid-}           &    827 & 16 & \texttt{Ġ(including}      &    769 \\
17 & \texttt{ĠtheĠuniversity}     &    784 & 17 & \texttt{ĠPeople's}        &    754 \\
18 & \texttt{ĠtheĠNorth}          &    758 & 18 & \texttt{ĠRomanĠEmpire}    &    751 \\
19 & \texttt{ĠtheĠChinese}        &    751 & 19 & \texttt{Ġd'}              &    719 \\
20 & \texttt{ĠannouncedĠthat}     &    749 & 20 & \texttt{thĠandĠ}          &    711 \\
\bottomrule
\end{tabular}
\caption{Top exclusive tokens by frequency on {SQuAD}.}
\label{tab:exclusive_squad}
\end{table*}

\begin{table*}[t]
\centering
\footnotesize
\setlength{\tabcolsep}{6pt}
\begin{tabular}{rlr@{\hskip 24pt}rlr}
\toprule
\multicolumn{3}{c}{\textbf{\myalgo only}} & \multicolumn{3}{c}{\textbf{SuperBPE only}} \\
\cmidrule(lr){1-3} \cmidrule(lr){4-6}
\textbf{Rank} & \textbf{Token} & \textbf{Frequency} & \textbf{Rank} & \textbf{Token} & \textbf{Frequency} \\
\midrule
1  & \texttt{S.}                       & 3{,}874 & 1  & \texttt{),Ġ}                 & 5{,}274 \\
2  & \texttt{ĠtheĠfourth}              & 1{,}637 & 2  & \texttt{UnitedĠStates}       & 4{,}956 \\
3  & \texttt{ĠtheĠlead}                & 1{,}260 & 3  & \texttt{U.S.}                & 3{,}864 \\
4  & \texttt{ĠunderĠtheĠageĠofĠ}       & 1{,}086 & 4  & \texttt{ĠonĠaĠ}              & 1{,}890 \\
5  & \texttt{ĠThereĠwere}              &    640 & 5  & \texttt{ĠfromĠtwo}           & 1{,}238 \\
6  & \texttt{ĠpeopleĠ(}                &    633 & 6  & \texttt{ĠwereĠ}              & 1{,}147 \\
7  & \texttt{ĠtheĠperiod}              &    632 & 7  & \texttt{ĠageĠ}               & 1{,}054 \\
8  & \texttt{InĠtheĠfirst}             &    599 & 8  & \texttt{ĠandĠaĠ}             & 1{,}020 \\
9  & \texttt{ĠtheĠloss}                &    592 & 9  & \texttt{ĠforĠaĠ}             &    996 \\
10 & \texttt{ĠHowĠmany}                &    516 & 10 & \texttt{ĠeveryĠ}             &    982 \\
11 & \texttt{ĠofĠindividuals}          &    450 & 11 & \texttt{ĠAfricanĠAmerican}   &    853 \\
12 & \texttt{.ĠInĠtheĠsecond}          &    437 & 12 & \texttt{'Ġ}                  &    775 \\
13 & \texttt{ĠaĠfemale}                &    415 & 13 & \texttt{ĠanĠ}                &    663 \\
14 & \texttt{Ġtogether,}               &    402 & 14 & \texttt{).ĠThere}            &    610 \\
15 & \texttt{ĠtheĠlongest}             &    401 & 15 & \texttt{Ġyears,Ġ}            &    583 \\
16 & \texttt{ĠearlyĠin}                &    394 & 16 & \texttt{ĠNativeĠAmerican}    &    561 \\
17 & \texttt{ĠfirstĠhalf}              &    366 & 17 & \texttt{Ġendzone}            &    462 \\
18 & \texttt{ĠtheĠsecondĠhalf}         &    330 & 18 & \texttt{-point}              &    401 \\
19 & \texttt{ĠtheĠSpanish}             &    308 & 19 & \texttt{-winning}            &    358 \\
20 & \texttt{ĠtheirĠhome}              &    300 & 20 & \texttt{ĠNewĠEngland}        &    314 \\
\bottomrule
\end{tabular}
\caption{Top exclusive tokens by frequency on {DROP}.}
\label{tab:exclusive_drop}
\end{table*}

\begin{table*}[t]
\centering
\footnotesize
\setlength{\tabcolsep}{6pt}
\begin{tabular}{rlr@{\hskip 24pt}rlr}
\toprule
\multicolumn{3}{c}{\textbf{\myalgo only}} & \multicolumn{3}{c}{\textbf{SuperBPE only}} \\
\cmidrule(lr){1-3} \cmidrule(lr){4-6}
\textbf{Rank} & \textbf{Token} & \textbf{Frequency} & \textbf{Rank} & \textbf{Token} & \textbf{Frequency} \\
\midrule
1  & \texttt{\_\{t}                              & 263 & 1  & \texttt{\}Ċ}                  & 1{,}483 \\
2  & \texttt{Ġ\textbackslash textbf\{}           & 219 & 2  & \texttt{\$Ċ}                  &    498 \\
3  & \texttt{\textbackslash end\{equation*\}}    & 205 & 3  & \texttt{*\}Ċ}                 &    409 \\
4  & \texttt{hat\{\textbackslash}                & 202 & 4  & \texttt{i,}                  &    294 \\
5  & \texttt{\textbackslash begin\{equation*\}}  & 198 & 5  & \texttt{]Ċ}                  &    288 \\
6  & \texttt{\},Ġ\textbackslash}                 & 185 & 6  & \texttt{,Ġ\textbackslash}    &    273 \\
7  & \texttt{Ġ\textbackslash text\{}             & 133 & 7  & \texttt{align}               &    256 \\
8  & \texttt{\textbackslash caption\{}           & 132 & 8  & \texttt{equation}            &    238 \\
9  & \texttt{hat\{}                              & 126 & 9  & \texttt{Ġ\textbackslash in}  &    231 \\
10 & \texttt{begin\{align*\}}                    & 126 & 10 & \texttt{\textbackslash in}   &    208 \\
11 & \texttt{end\{align*\}}                      & 126 & 11 & \texttt{\textbackslash begin\{equation} & 198 \\
12 & \texttt{\_\{k}                              & 122 & 12 & \texttt{\textbackslash item} &    175 \\
13 & \texttt{boldsymbol\{\textbackslash}         & 103 & 13 & \texttt{.Ċ}                  &    164 \\
14 & \texttt{\{x}                                & 101 & 14 & \texttt{\$.Ċ}                &    140 \\
15 & \texttt{mathbf\{x}                          &  89 & 15 & \texttt{Ġ\textbackslash \textbackslash}Ċ & 134 \\
16 & \texttt{tilde\{}                            &  84 & 16 & \texttt{left(}               &    121 \\
17 & \texttt{\_\{\textbackslash text\{}          &  82 & 17 & \texttt{(t}                  &    116 \\
18 & \texttt{\}Ġ=}                               &  82 & 18 & \texttt{Ġ+Ġ\textbackslash}   &    115 \\
19 & \texttt{\}Ġ=Ġ\textbackslash}                &  81 & 19 & \texttt{\textbackslash \textbackslash}Ċ &    112 \\
20 & \texttt{theta\_}                            &  79 & 20 & \texttt{left(\textbackslash} &    101 \\
\bottomrule
\end{tabular}
\caption{Top exclusive tokens by frequency on {Image2Struct}.}
\label{tab:exclusive_img2struct}
\end{table*}

\section*{Appendix G:  Details on Sensitivity Analysis}
\label{app:sensitivity}
Table~\ref{tab:ael_settings} reports the AEL obtained on the evaluation data when each parameter is varied within the bounds from the previous section, while keeping all other settings fixed. The first row corresponds to the configuration used in the main experiments ($\theta_{\mathtt{general}}^{\mathtt{min}} = 0.3$, $\theta_{\mathtt{latex}}^{\mathtt{min}} = 0.08$); the remaining rows progressively relax the thresholds.

\begin{table*}[t!]
\centering

\setlength{\tabcolsep}{3pt} 
\renewcommand{\arraystretch}{0.9}  
\scalebox{0.8}{
\begin{tabular}{l c c c c c c}
\toprule
\textbf{Parameters} & \textbf{$|V|$} & \textbf{DROP} & \textbf{SQuAD} & \textbf{CSQA} & \textbf{LS} & \textbf{I2S} \\
\midrule
\textit{baseline} & 204,601 & 210.43 & 126.03 & 12.24 & 54.88 & 207.96 \\
\midrule
$\theta_{latex}^{min}=0.05,$ & 181,752 & 222.60 & 136.02 & 13.44 & 52.93 & 202.47 \\
$\theta_{general}^{min}\!=\!0.5$  & & \textit{(-5.78\%)} & \textit{(-7.92\%)} & \textit{(-9.80\%)} & \textit{(+3.55\%)} & \textit{(+2.64\%)} \\
\cmidrule{1-7}
$\theta_{latex}^{min}=0.08$, & 168,292 & 222.60 & 136.02 & 13.44 & 55.75 & 211.63 \\
 $\theta_{general}^{min}\!=\!0.5$ & & \textit{(-5.78\%)} & \textit{(-7.92\%)} & \textit{(-9.80\%)} & \textit{(-1.58\%)} & \textit{(-1.76\%)} \\
\cmidrule{1-7}
$\theta_{latex}^{min}=0.05,$  & 208,043 & 210.43 & 126.03 & 12.24 & 52.57 & 200.41 \\
 $\theta_{general}^{min}\!=\!0.3$ & & \textit{(+0.0\%)} & \textit{(+0.0\%)} & \textit{(+0.0\%)} & \textit{(+4.20\%)} & \textit{(+3.63\%)} \\
\cmidrule{1-7}
$\theta_{latex}^{min}=0.08,$ & 247,260 & 208.21 & 124.25 & 12.04 & 54.86 & 207.90 \\
$\theta_{general}^{min}\!=\!0.25$ & & \textit{(+1.05\%)} & \textit{(+1.41\%)} & \textit{(+1.63\%)} & \textit{(+0.03\%)} & \textit{(+0.02\%)} \\
\cmidrule{1-7}
$\theta_{latex}^{min}=0.05,$  & 261,512 & \textbf{206.33} & \textbf{124.24} & \textbf{12.04} & \textbf{52.11} & \textbf{198.89} \\
 $\theta_{general}^{min}\!=\!0.25$ & & \textbf{(+1.94\%)} & \textbf{(+1.42\%)} & \textbf{(+1.63\%)} & \textbf{(+5.04\%)} & \textbf{(+4.36\%)} \\
\cmidrule{1-7}
$\theta_{latex}^{min}=0.05,$   & 317,155 & 205.08 & 122.57 & 11.86 & 51.86 & 197.48 \\
$\theta_{general}^{min}\!=\!0.23$ & & \textit{(+2.60\%) }& \textit{(+2.74\%)} & \textit{(+3.10\%) }& \textit{(-5.50\%)} & \textit{(+5.03\%)} \\
\cmidrule{1-7}
$\theta_{latex}^{min}=0.05,$   & 646,506 & 201.14 & 119.65 & 11.51 & 51.13 & 196.25 \\
$\theta_{general}^{min}\!=\!0.2$ & & \textit{(+4.41\%)} & \textit{(+5.06\%)} & \textit{(+5.96\%)} & (+6.83\%) & \textit{(+5.63\%)} \\
\cmidrule{1-7}
$\theta_{latex}^{min}=0.08,$   & 1,685,088 & 183.60 & 107.57 & 10.09 & 54.93 & 204.55 \\
$\theta_{general}^{min}\!=\!0.1$ & & \textit{(+12.75\%)} & \textit{(+14.64\%)} & \textit{(+17.56\%)} & \textit{(-0.09\%)} & \textit{(+1.63\%)} \\
\bottomrule
\end{tabular}
}
\caption{Sensitivity analysis of \myalgo: AEL scores for  different settings of $\theta_{general}^{min}$ and $\theta_{latex}^{min}$. Values in parentheses indicate the percentage variation relative to the baseline.}
\label{tab:ael_settings}
\end{table*}

\end{document}